\def\year{2021}\relax
\documentclass[letterpaper]{article} 
\usepackage{aaai21}  
\usepackage{times}  
\usepackage{helvet} 
\usepackage{courier}  
\usepackage[hyphens]{url}  
\usepackage{graphicx} 
\usepackage{graphicx}  
\author{Amirreza Farnoosh, Bahar Azari, Sarah Ostadabbas\\ Department of Electrical and Computer
Engineering, Northeastern University,\\ Boston, Massachusetts,
USA}

\usepackage[utf8]{inputenc} 
\usepackage[T1]{fontenc}    
\usepackage{hyperref}       
\usepackage{url}            
\usepackage{booktabs}       
\usepackage{amsfonts}       
\usepackage{nicefrac}       
\usepackage{microtype}      
\usepackage{caption}
\usepackage{subcaption}
\usepackage{graphicx}
\usepackage{amsmath}
\usepackage{amssymb}
\usepackage{amsthm}
\usepackage{color}
\usepackage{soul}

\usepackage{natbib}
\usepackage{bm}
\usepackage{relsize}
\usepackage{array}
\usepackage{diagbox}
\usepackage{multirow}
\usepackage{hyperref}
\usepackage{relsize}
\definecolor{cobalt}{rgb}{0.0, 0.28, 0.67}

\newcommand{\eqnref}[1]{Eq.~(\ref{eqn:#1})}
\newcommand{\figref}[1]{Fig.~\ref{fig:#1}}
\newcommand{\tblref}[1]{Table~\ref{tbl:#1}}
\newcommand{\secref}[1]{Section~\ref{sec:#1}}

\newcommand{\suppref}[1]{Appendix~\ref{app:#1}}

\newcommand{\graphicalmodel}{
\begin{figure*}[t]
\centering
\begin{subfigure}[b]{.43\textwidth}
\includegraphics[width=\linewidth]{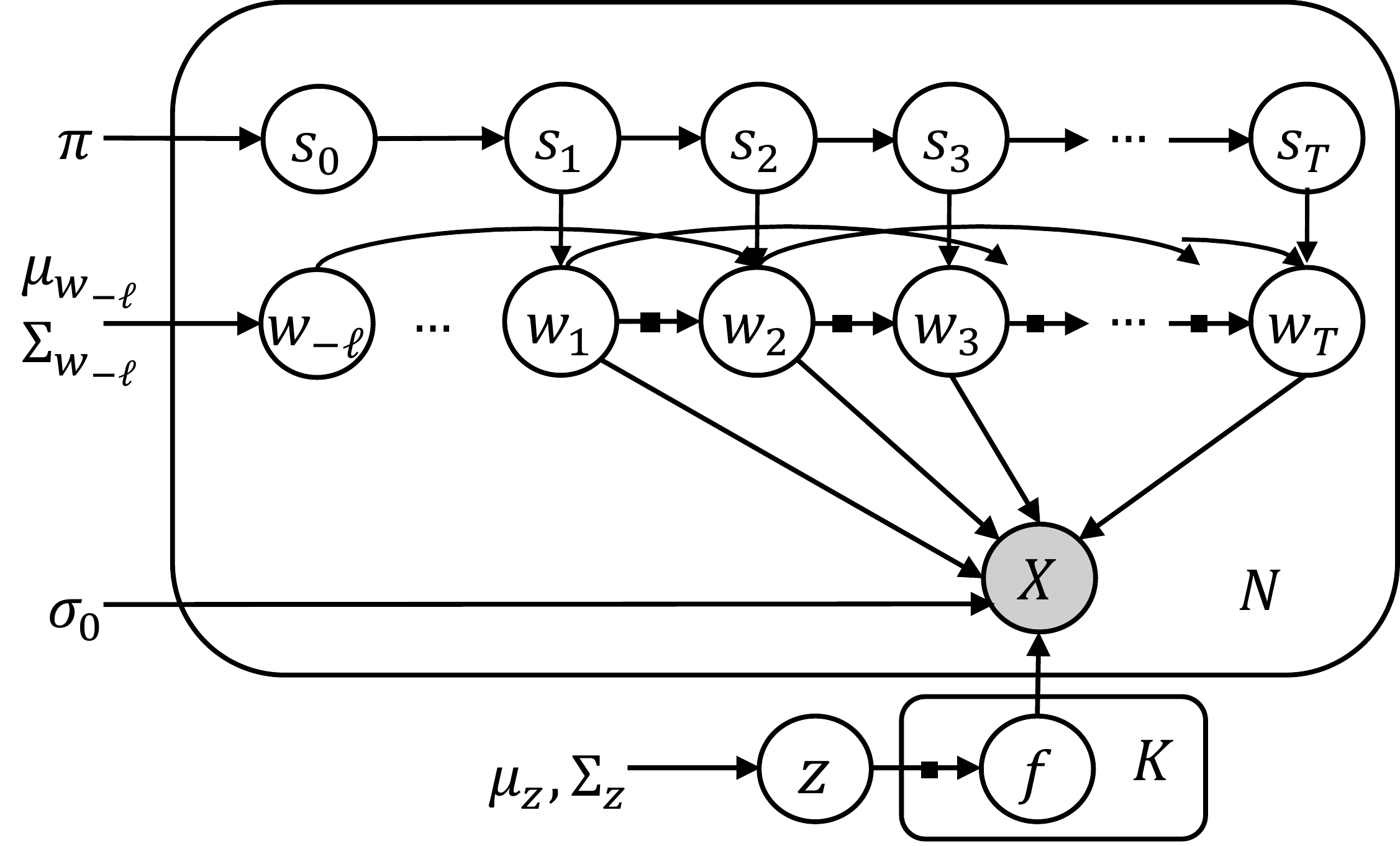}
\end{subfigure}
\begin{subfigure}[b]{.56\textwidth}
\includegraphics[width=1\linewidth]{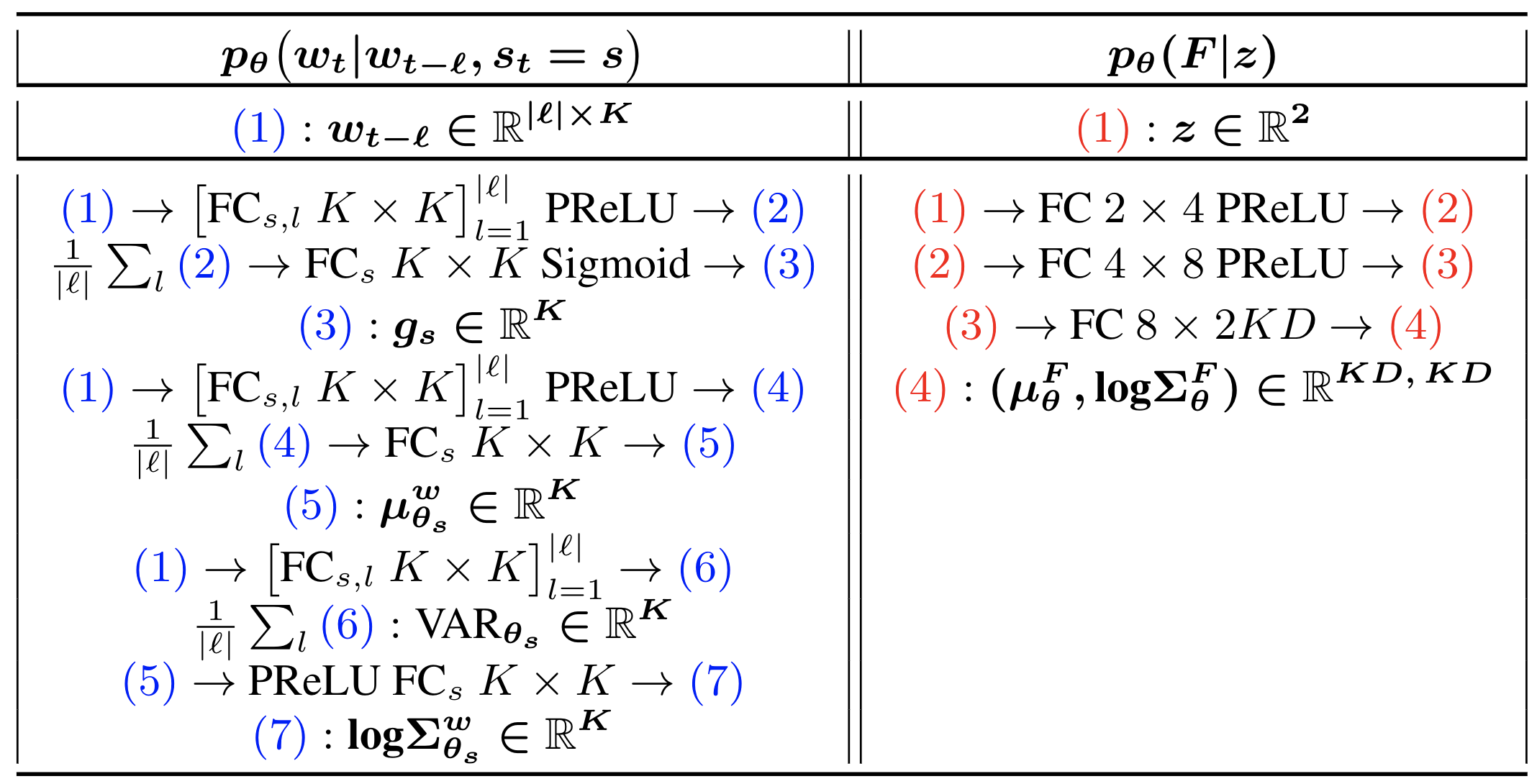}
\end{subfigure}
 \caption{\small \textbf{Left}: Graphical model representation for deep switching auto-regressive factorization (DSARF). Temporal weights, $w_{1:T}$ are generated according to a nonlinear auto-regressive model, \emph{switched} by a Markovian chain of discrete states, $s_{1:t}$. Spatial factors, $f_{1:K}$, come from a shared low-dimensional latent, $z$. The solid black squares represent nonlinear functions. \textbf{Right}: Network architectures parameterizing the nonlinear mappings employed in DSARF. A fully connected (FC) layer is defined for each state $s\in\{1,\dots,S\}$, and lag $l\in\ell$ as FC$_{s,l}$. These layers take as input $w_{t-\ell}$, and their outputs are aggregated in the succeeding layer: e.g., $\boldsymbol{\mu_{\theta_s}^w}=\text{FC}_s(\frac{1}{|\ell|}\sum_{l\in\ell} \text{PReLU}(\text{FC}_{s,l}(w_{t-l})))$.}
 \label{fig:graph}
\end{figure*}
}

\newcommand{\synthfig}{
\begin{figure*}[t]
\centering
\begin{subfigure}[b]{0.235\linewidth}
\includegraphics[width=1.0\linewidth]{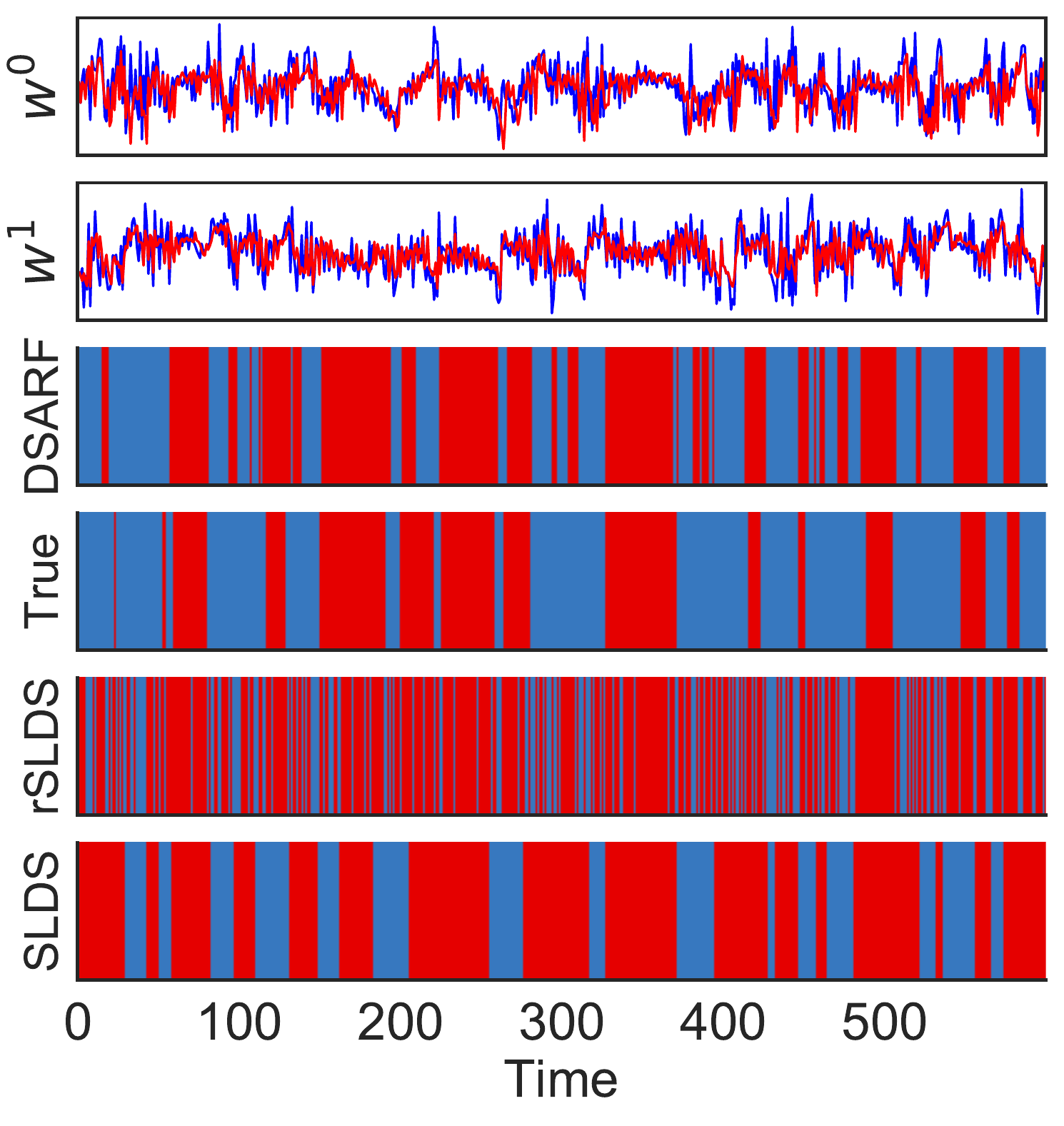}
\caption{Toy example}
\label{fig:toy}
\end{subfigure}
\begin{subfigure}[b]{0.39\linewidth}
\begin{subfigure}[b]{0.6\linewidth}
\includegraphics[width=1.0\linewidth]{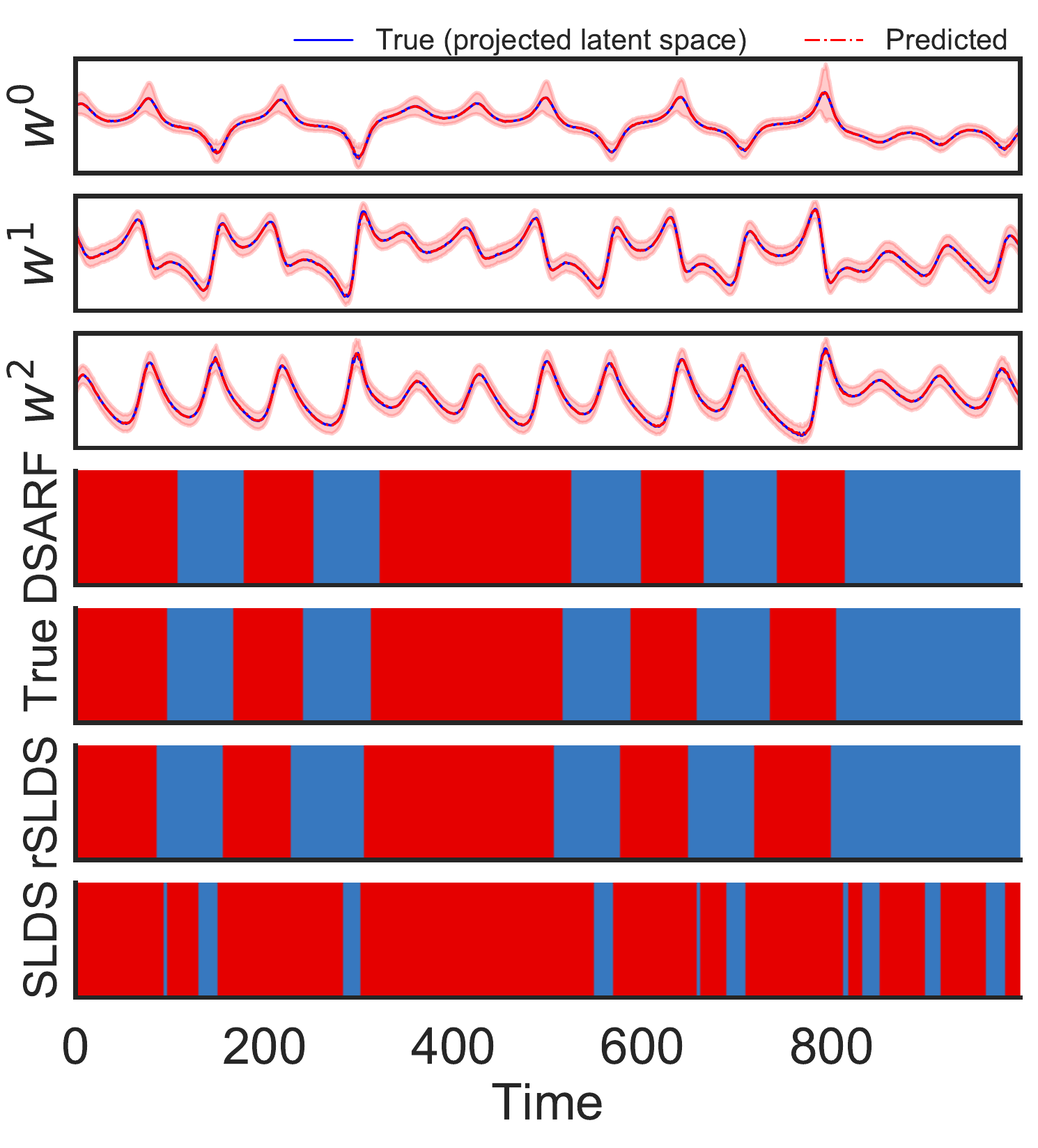}
\end{subfigure}
\begin{subfigure}[b]{0.32\linewidth}
\includegraphics[width=1.0\linewidth]{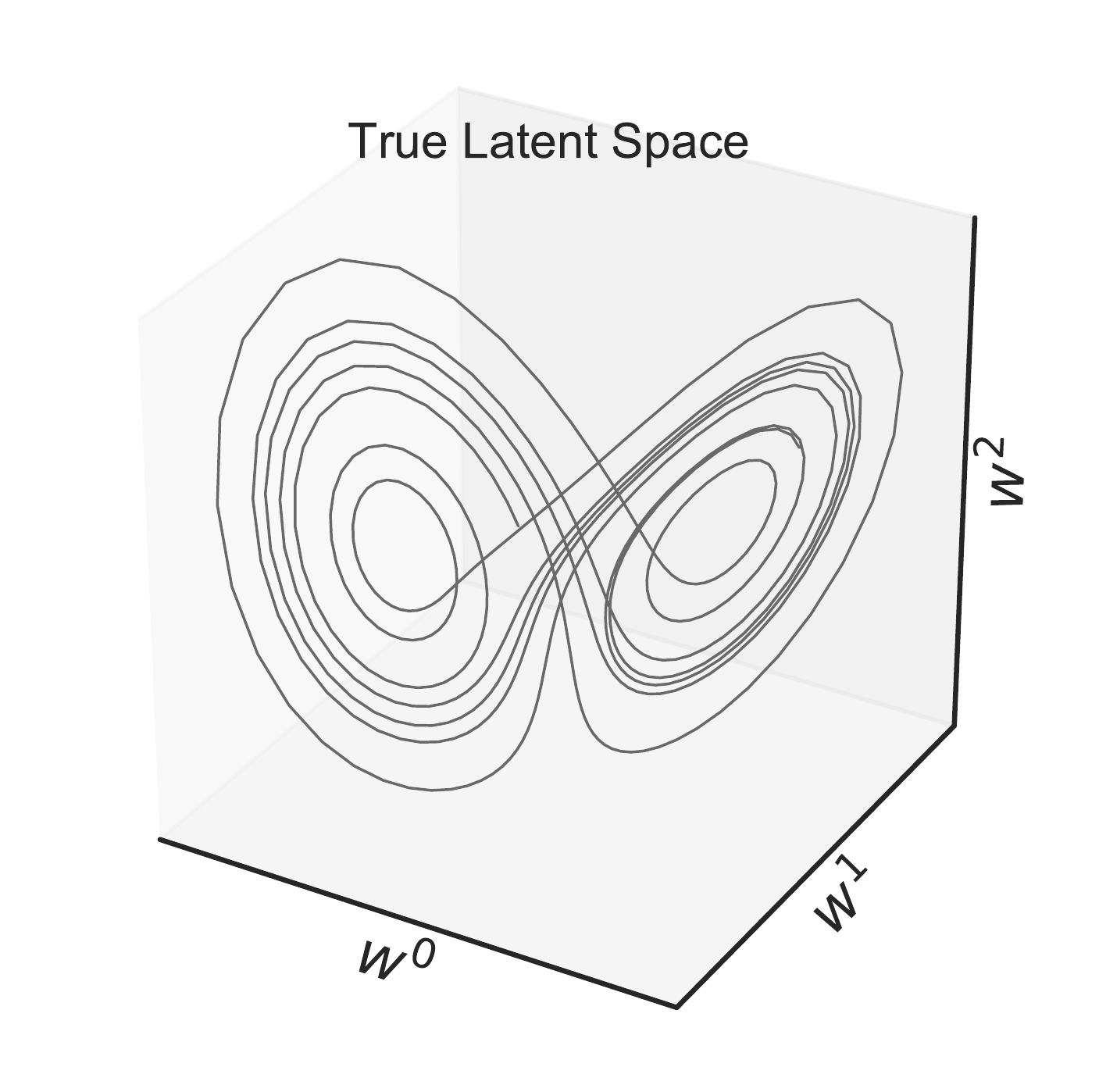}\\
\includegraphics[width=1.0\linewidth]{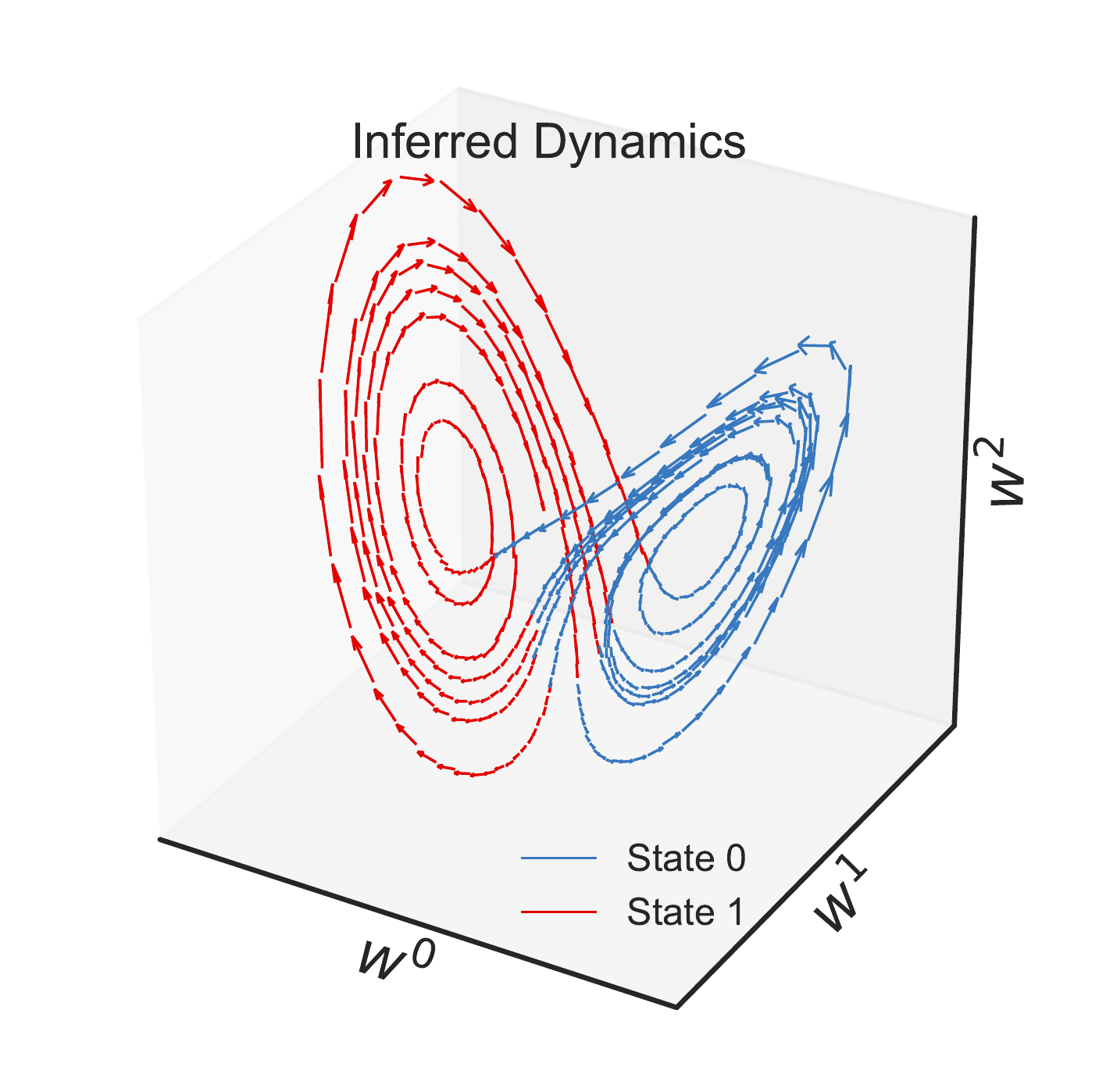}
\end{subfigure}
\caption{Lorenz attractor}
\end{subfigure}
\hspace{-4mm}
\begin{subfigure}[b]{0.36\linewidth}
\begin{subfigure}[b]{0.64\linewidth}
\includegraphics[width=1\linewidth]{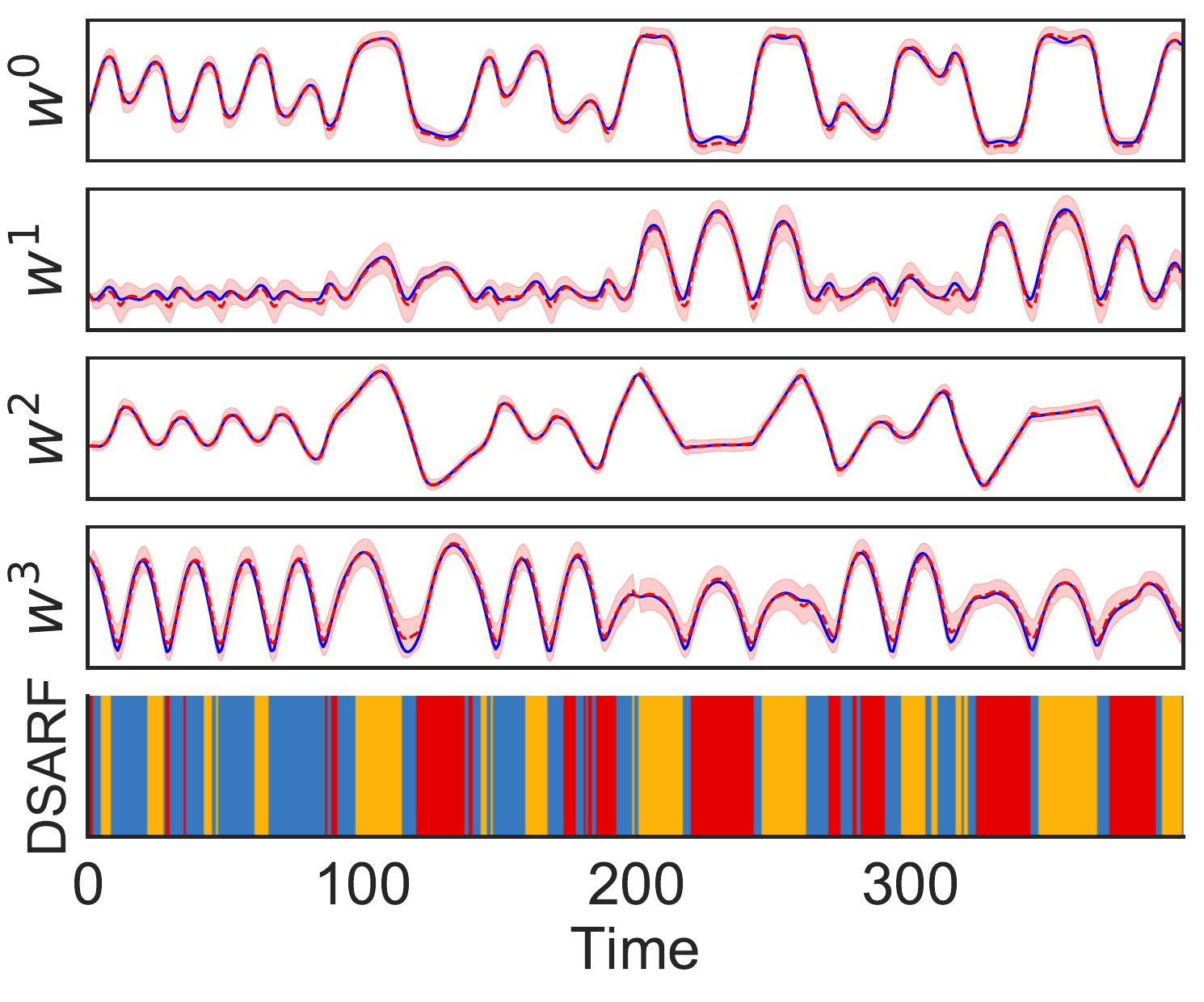}
\vspace{0.2mm}
\end{subfigure}
\begin{subfigure}[b]{0.3\linewidth}
\includegraphics[width=1.2\linewidth]{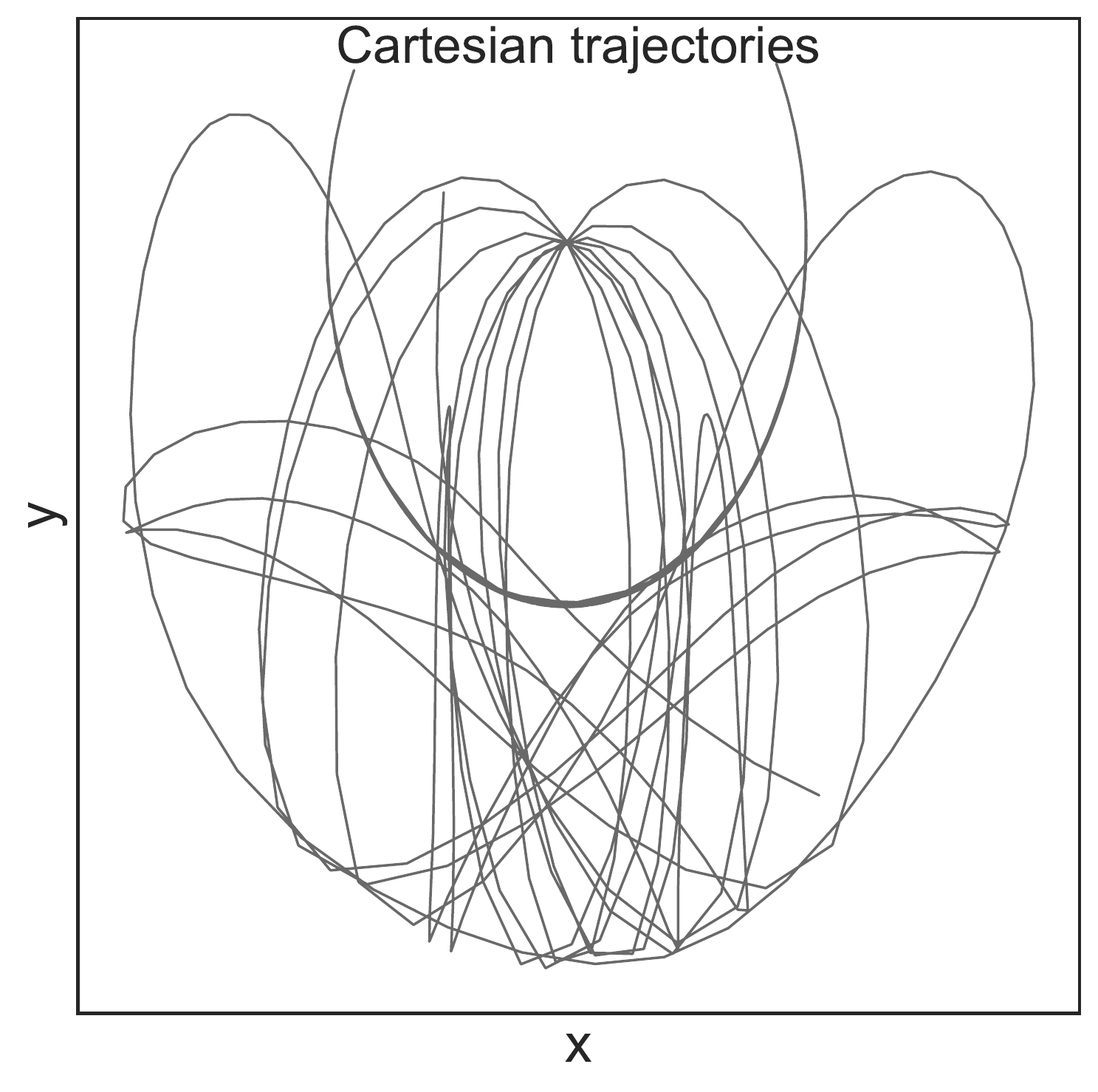}\\
\includegraphics[width=1.2\linewidth]{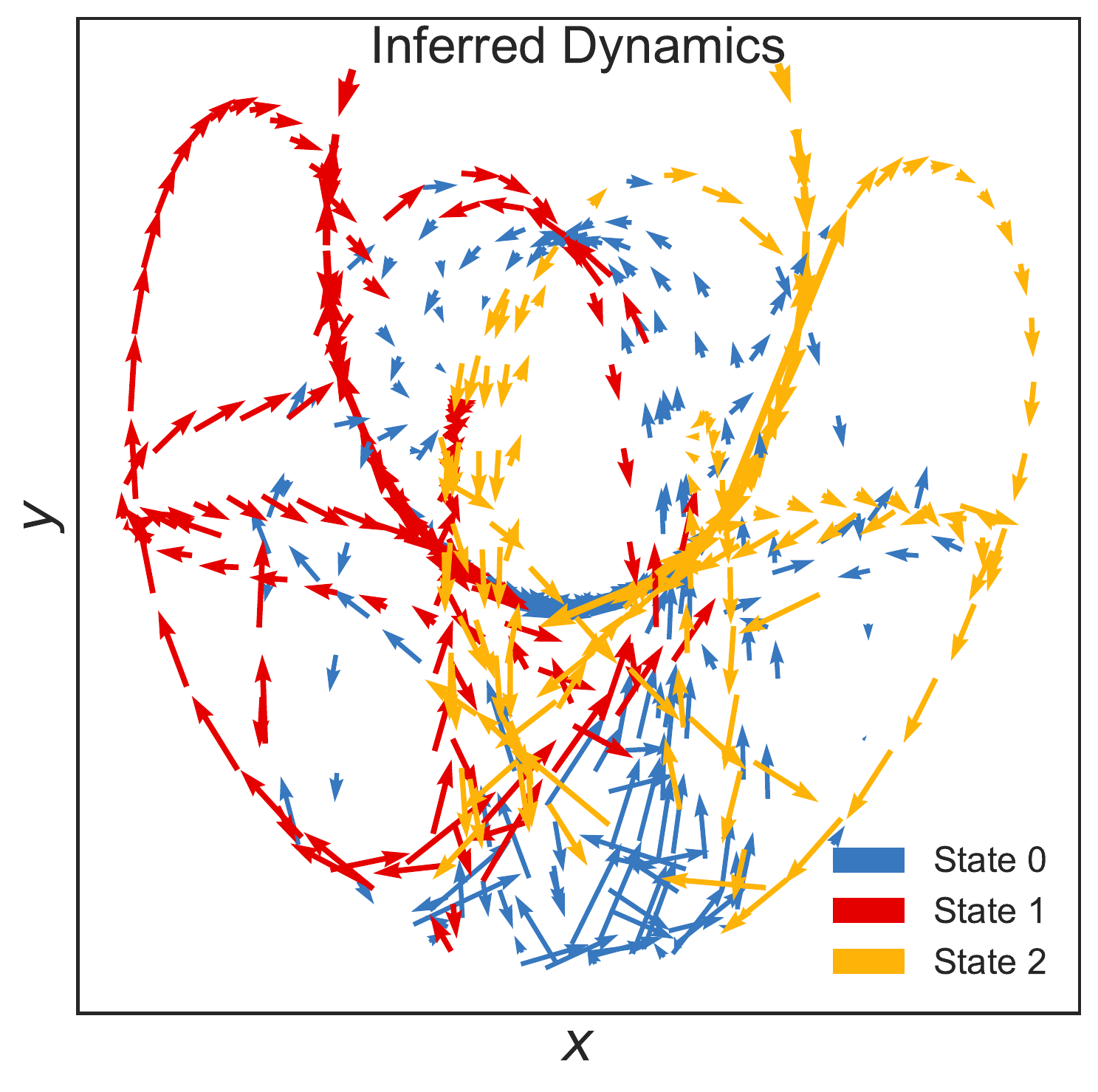}
\end{subfigure}
\caption{Double pendulum}
\end{subfigure}
\caption{{\small Test set results for synthetic experiments. (a) DSARF with $\ell=\{1,2,3\}$ outperforms rSLDS and SLDS in recovering the actual temporal states as these baselines failed in modeling the higher order temporal dependencies in this synthetic data. (b) DSARF separates the two planes in lorenz attractor into two distinct states each with rotational dynamics similar to rSLDS, while SLDS completely failed in this task. (c) DSARF with $\ell=\{1,2\}$ outperforms baselines in short-term prediction. This is expected as the motion of pendulums are governed by a set of coupled second-order ordinary differential equations. We observe that for $S=3$ the dynamical trajectory is roughly segmented along the deflection angle of the first pendulum. We have visualized the true (blue) and predicted (red) latent space for all the experiments (see \suppref{synthetic_details} for more details). Red shaded regions correspond to prediction uncertainty.}}
\label{fig:synthetic}    
\end{figure*}
}
\newcommand{\longfig}{
\begin{figure*}[t]
\centering
    \begin{subfigure}[b]{0.3\textwidth}
         \centering
         \includegraphics[width=1\textwidth]{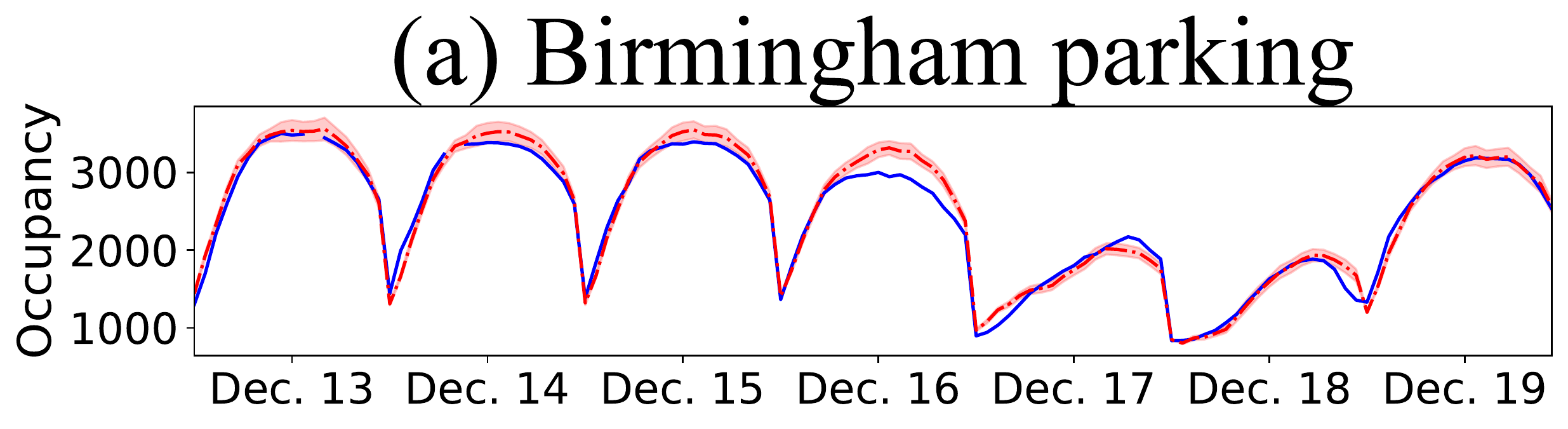}
         \label{fig:bir_long}
     \end{subfigure}
     \hfill
     \begin{subfigure}[b]{0.3\textwidth}
         \centering
         \includegraphics[width=1\textwidth]{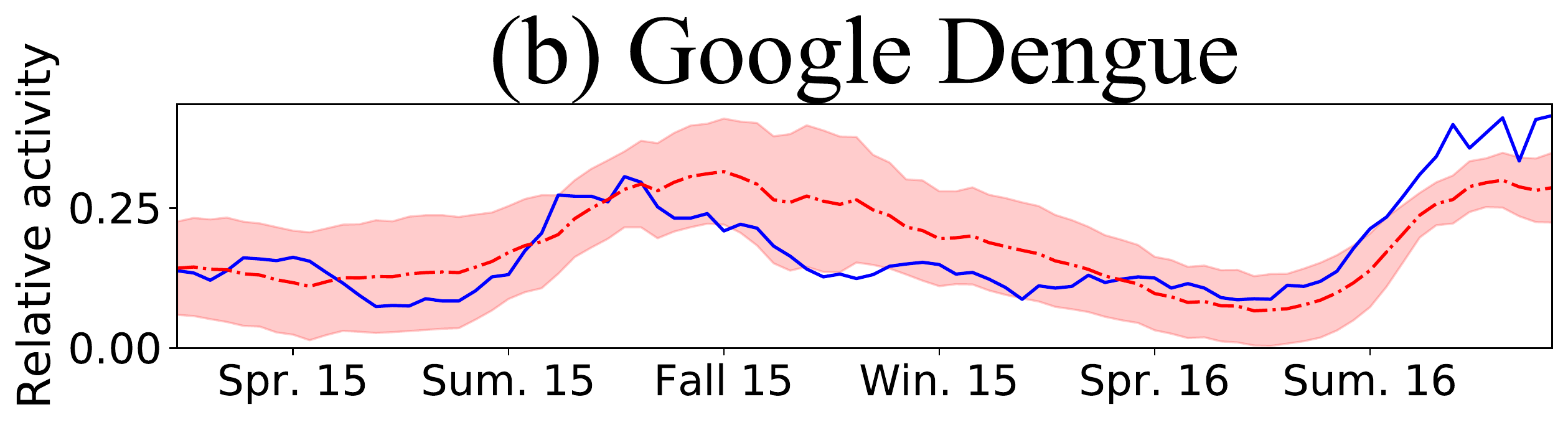}
         \label{fig:deng_long}
     \end{subfigure}
     \hfill
     \begin{subfigure}[b]{0.3\textwidth}
         \centering
         \includegraphics[width=1\textwidth]{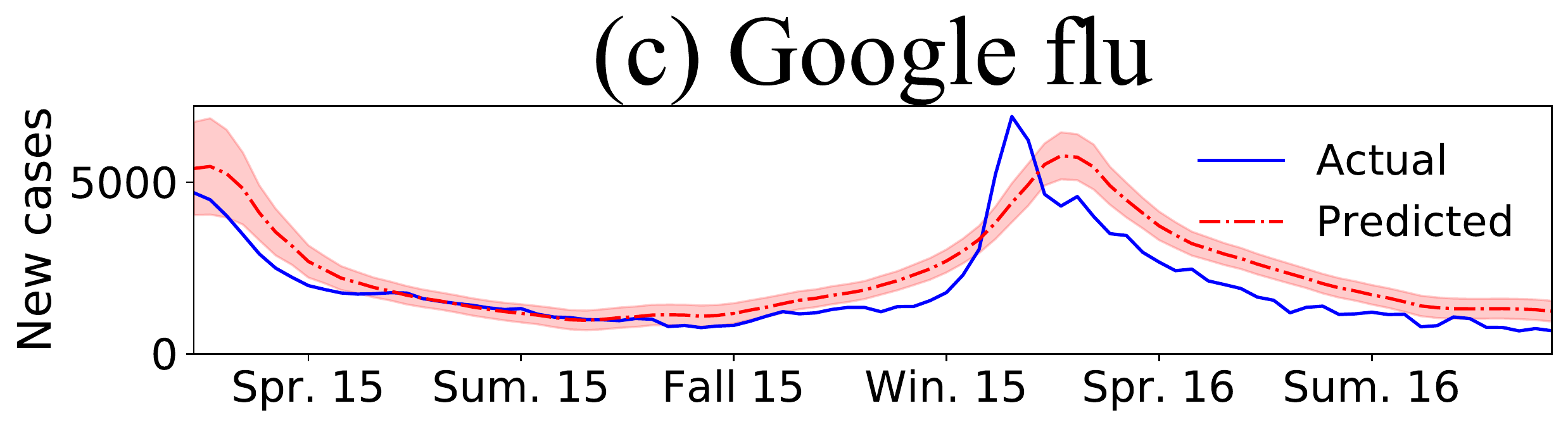}
         \label{fig:flu_long}
     \end{subfigure}
     \\ 
     \vspace{-4mm}
    \begin{subfigure}[b]{0.3\textwidth}
        \centering
        \includegraphics[width=1\textwidth]{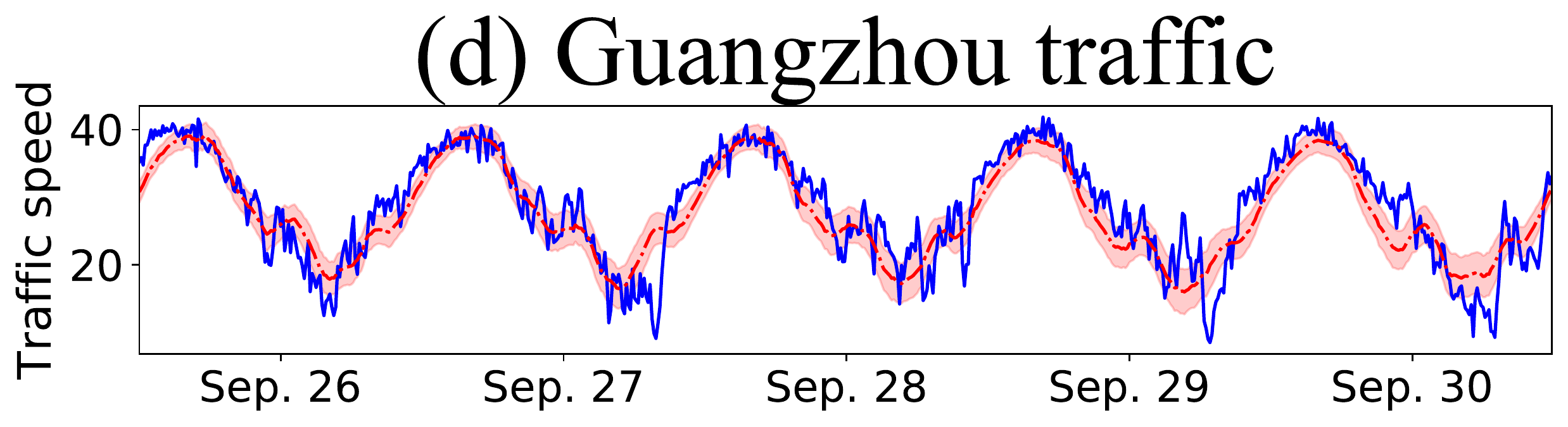}
        \label{fig:guan_long}
    \end{subfigure}
    \hfill
    \begin{subfigure}[b]{0.3\textwidth}
        \centering
        \includegraphics[width=1\textwidth]{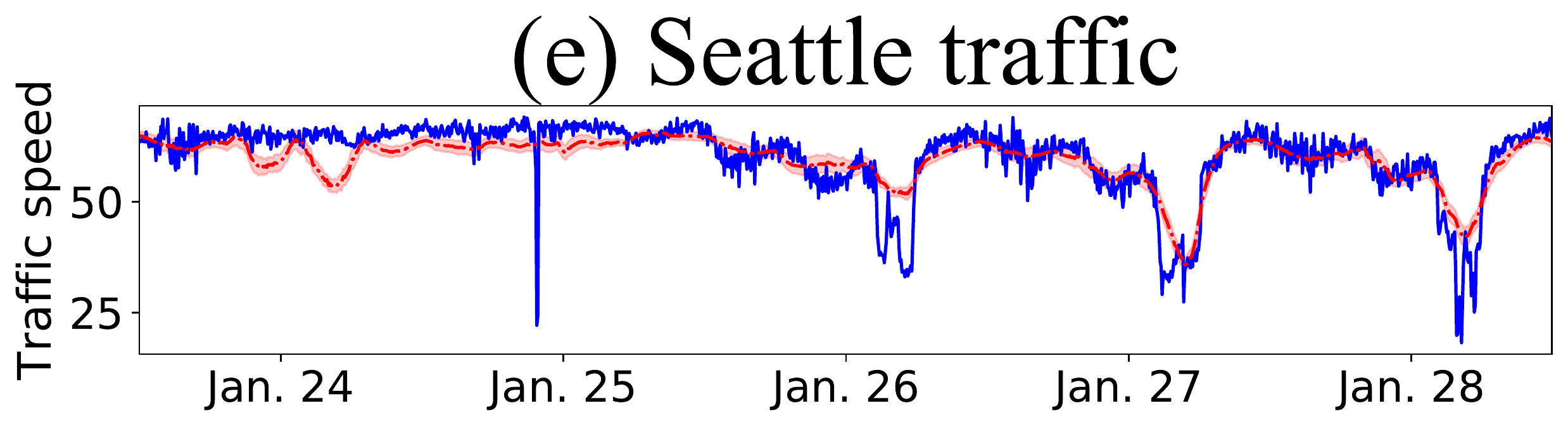}
        \label{fig:sea_short}
    \end{subfigure}
    \hfill
    \begin{subfigure}[b]{0.3\textwidth}
        \centering
        \includegraphics[width=1\textwidth]{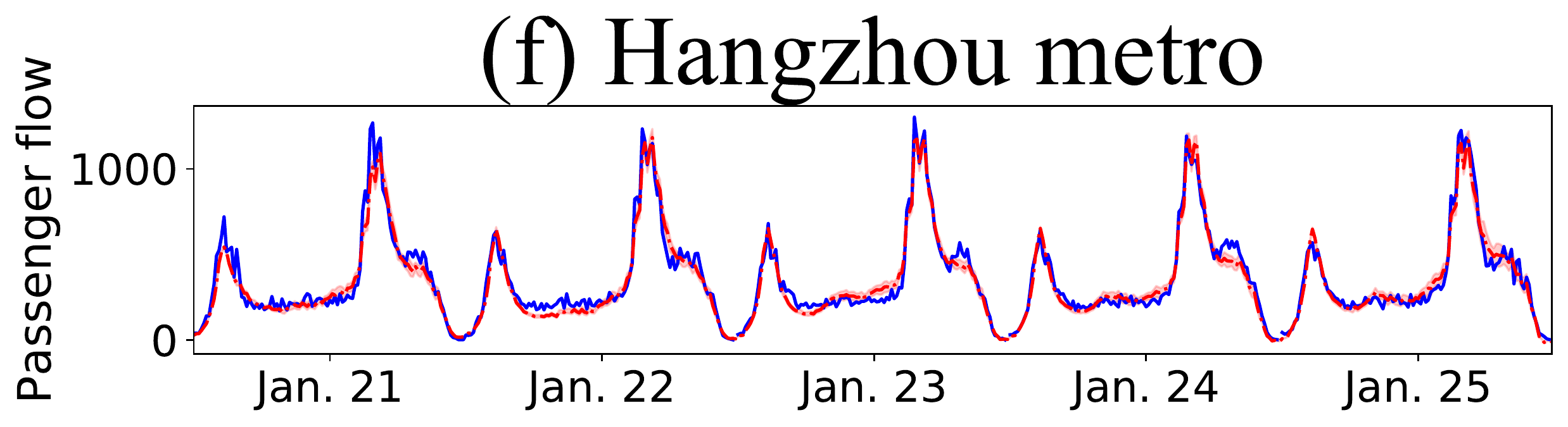}
        \label{fig:han_long}
    \end{subfigure}
    \\ 
    \vspace{-4mm}
    \begin{subfigure}[b]{0.3\textwidth}
        \centering
        \begin{subfigure}[b]{\textwidth}
        \centering
        \includegraphics[width=1\textwidth]{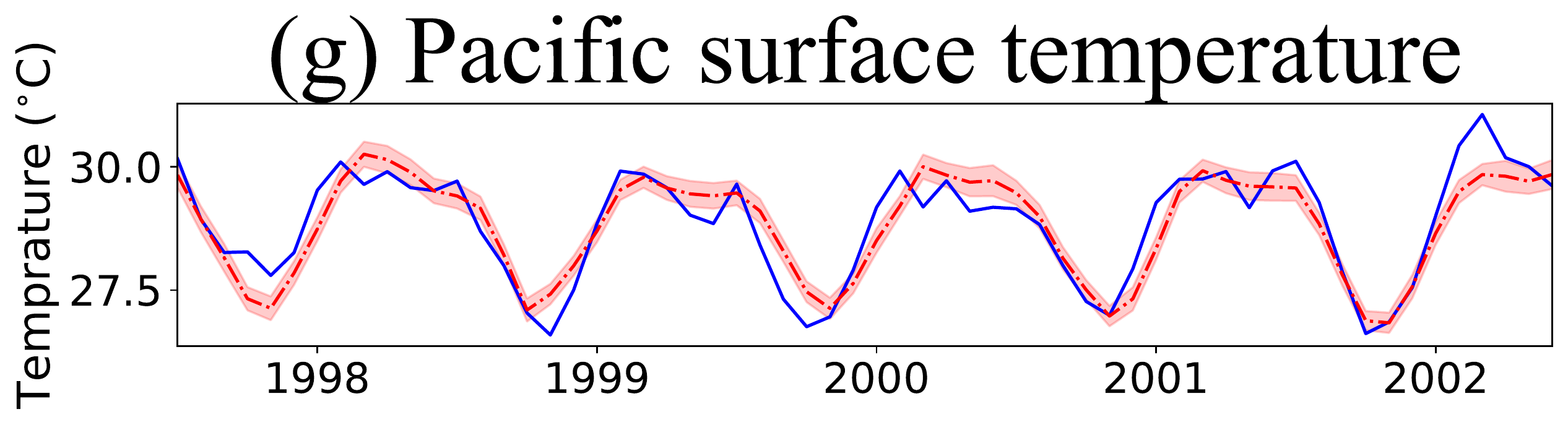}
        \label{fig:pac_long}
        \end{subfigure}
        \\
        \vspace{-4mm}
        \begin{subfigure}[b]{\textwidth}
        \centering
        \includegraphics[width=1\textwidth]{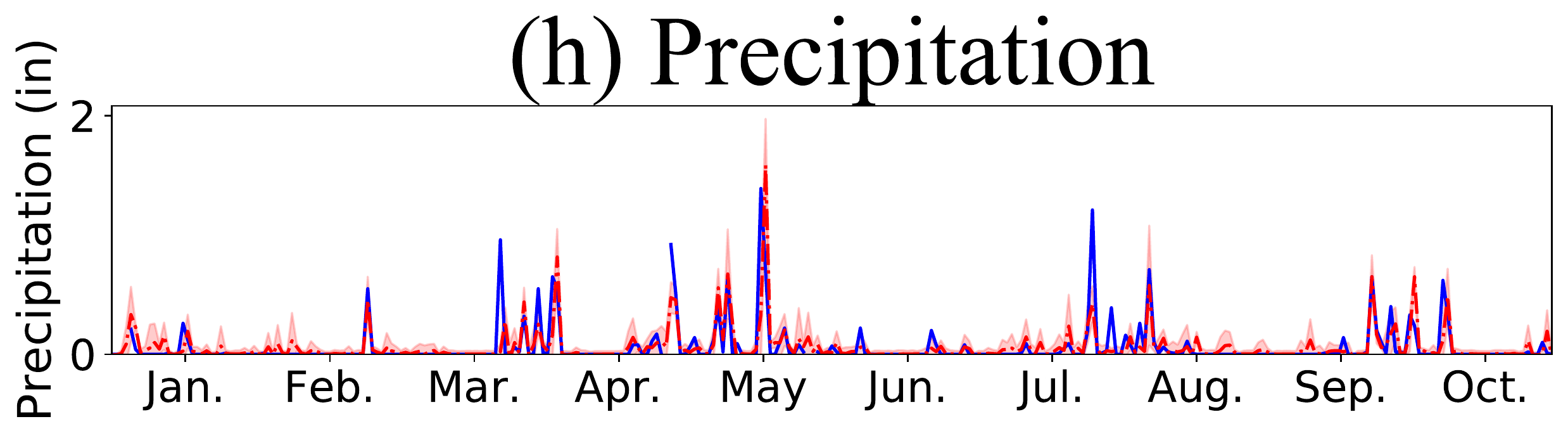}
        \label{fig:prec_short}
        \end{subfigure}
    \end{subfigure}
    \hfill
    \begin{subfigure}[b]{0.3\textwidth}
    \centering
    \includegraphics[width=1\textwidth]{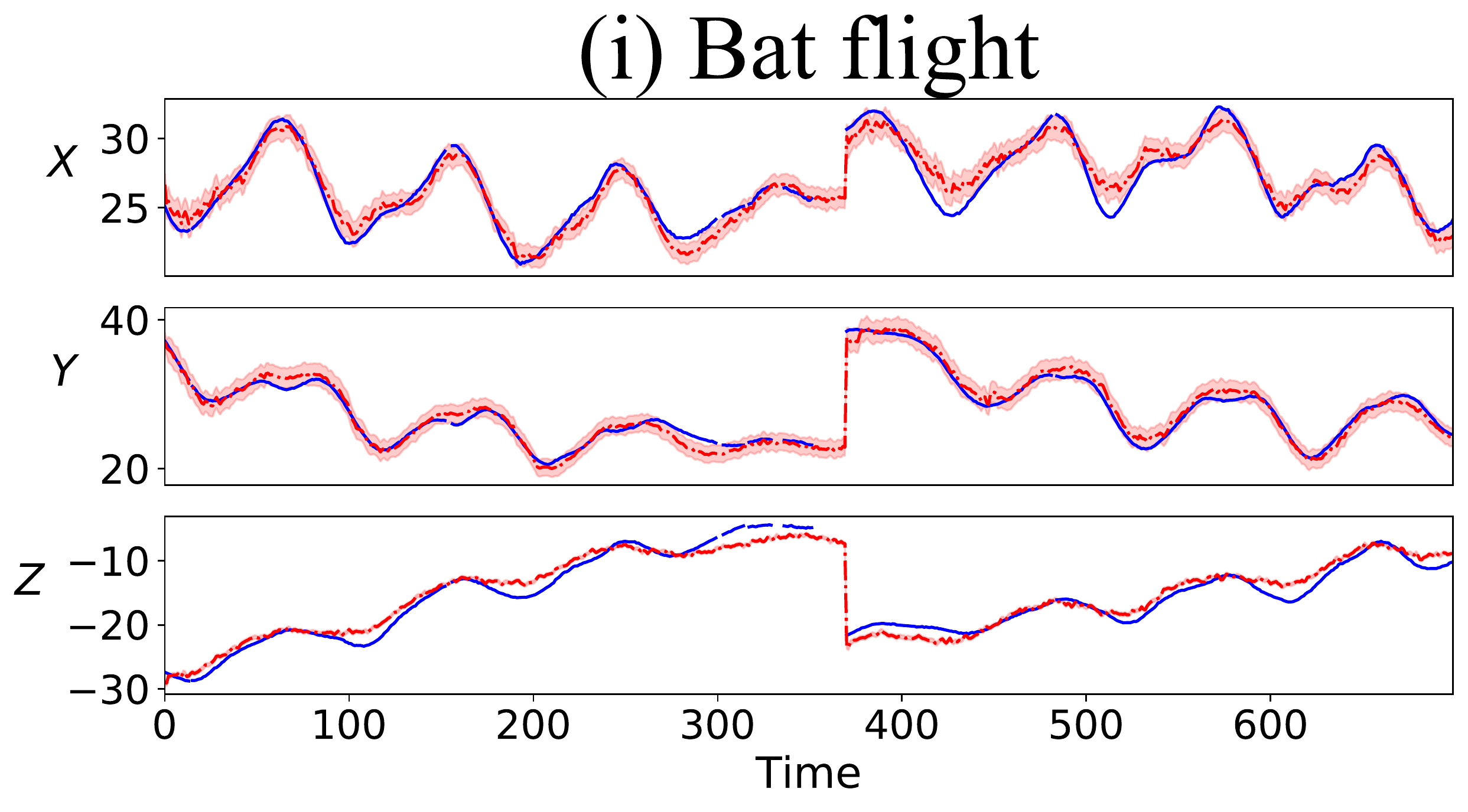}
    \label{fig:bat_short}
    \end{subfigure}
    \hfill
    \begin{subfigure}[b]{0.3\textwidth}
    \centering
    \includegraphics[width=1\textwidth]{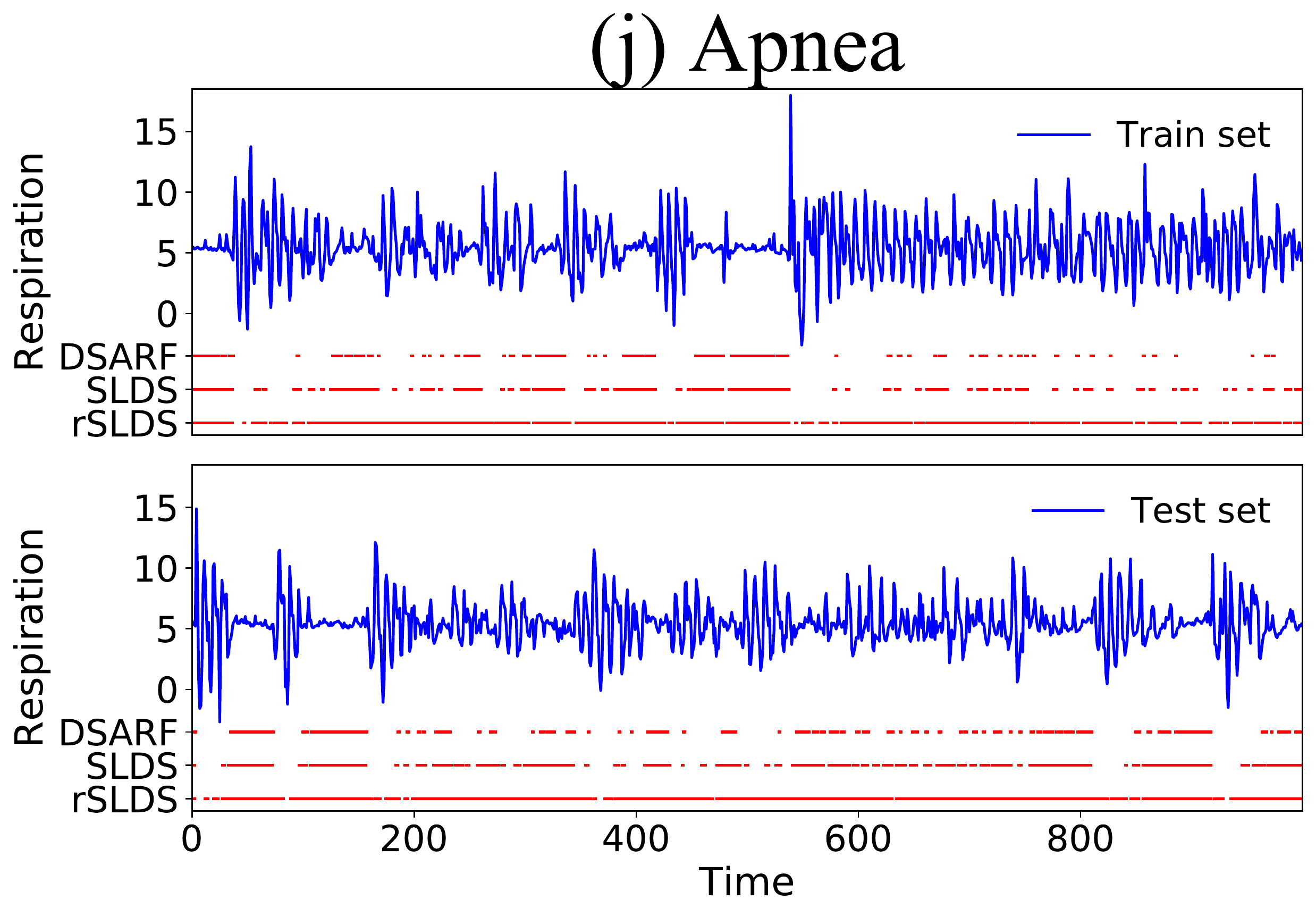}
    \label{fig:apnea_long}
    \end{subfigure}
    \caption{{\small \textbf{(a)-(g)}: Long-term predictions of test sets for real-world datasets. \textbf{(h), (i)}: Short-term predictions of test sets for Precipitation and Bat flight datasets. Red shaded regions correspond to prediction uncertainty. One spatial dimension per dataset is visualized. \textbf{(j)}: DSARF segments the respiration signal for both train and test sets into instances of apnea, outperforming SLDS and rSLDS.}}
    \label{fig:long}    
\end{figure*}
}

\newcommand{\penfigsupp}{
\begin{figure}[t]
 \centering
 \includegraphics[width=0.35\linewidth]{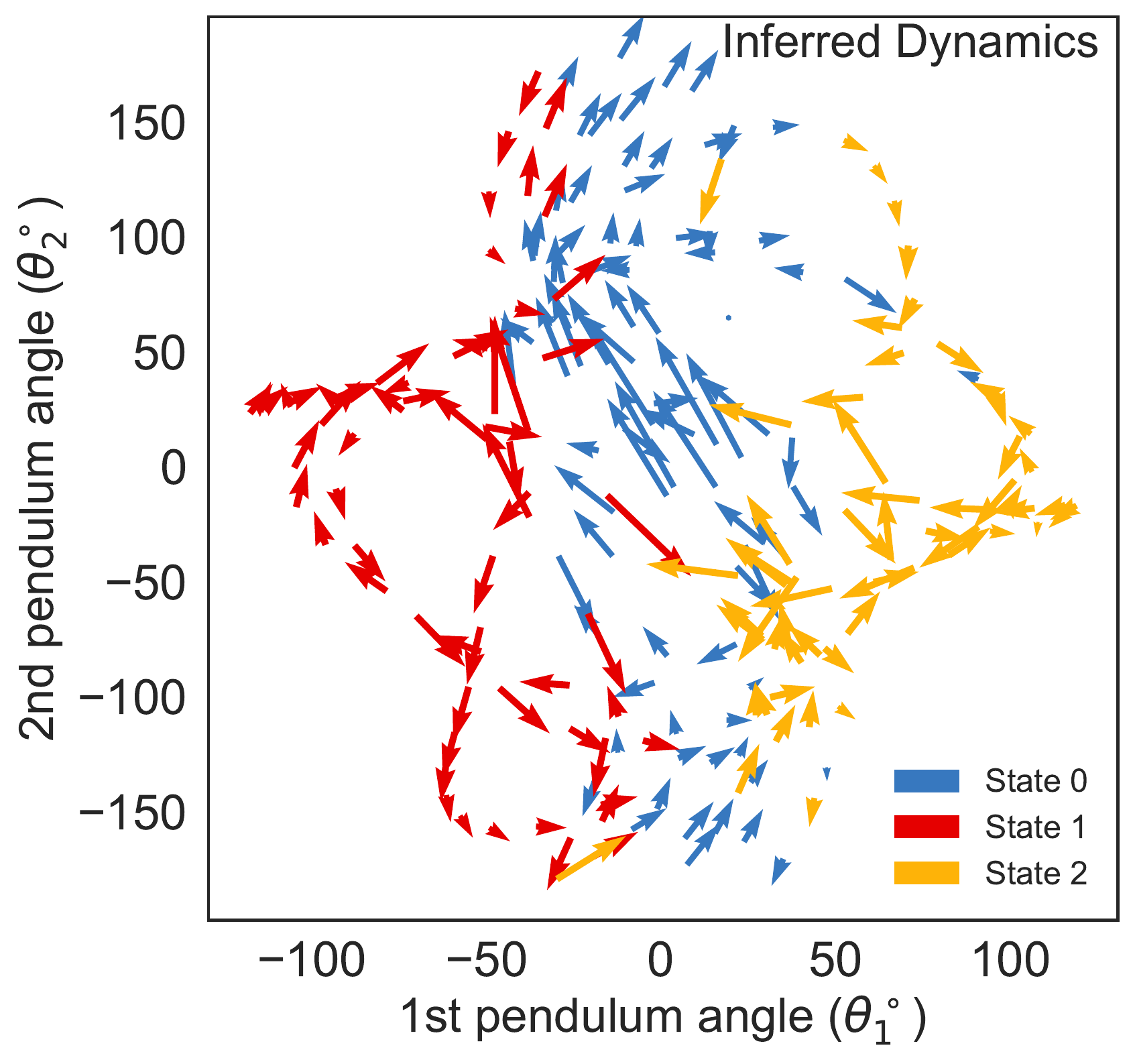}
 \caption{\small{Inferred dynamical trajectory with $S=3$ for the double pendulum experiment in $\theta_1$--$\theta_2$ space. The dynamical space is roughly segmented along $\theta_1$, the deflection angle of the first pendulum.}}
 \label{fig:penfigsupp}
\end{figure}
}

\newcommand{\comparisontab}{
\begin{table*}[!t]
{\centering
\caption{Performance comparison of short-time and long-term prediction. DMSTF outperforms on all datasets, doing significantly better particularly on the Birmingham dataset.}
\label{tab:pre}
\scriptsize
\begin{center}
\begin{tabular}{l|cc|cc|cc|cc}
\toprule
\multirow{2}{*}{\backslashbox{Dataset}{Model}} &  \multicolumn{2}{c|}{DMSTF} & \multicolumn{2}{c|}{BTMF} & \multicolumn{2}{c|}{BayesTRMF}& \multicolumn{2}{c}{TRMF} \\
& \tiny{NRMSE(\%)}   & \tiny{MAPE(\%)}    & \tiny{NRMSE(\%)}   & \tiny{MAPE(\%)}   & \tiny{NRMSE(\%)}   & \tiny{MAPE(\%)}      & \tiny{NRMSE(\%)}   & \tiny{MAPE(\%)}  \\
\midrule
Birmingham &  \textbf{5.76} & \textbf{17.30}   & 15.27 & 25.10 &  15.84 & 31.80   & 17.13  & 32.63\\
\midrule
Guangzhou &  \textbf{10.20} & \textbf{10.13}   & 10.30  & 10.25 &  10.75 & 10.70   & 10.83 & 10.65\\
\midrule
Hangzhou &  \textbf{15.55}  &  29.73   & 16.67  & 30.04 &  18.26 & 30.17   & 17.86  & \textbf{27.77}\\
\midrule
Seattle &  \textbf{7.52}  &  \textbf{7.27}   & 7.69  & 7.48 &  8.10 & 7.90   & 8.30  & 7.96\\
\midrule
Bat &  \textbf{8.08}  &  \textbf{7.39}   & 8.89  &  &  8.16 &    & 9.02  & \\
\midrule
PST &  \textbf{2.53}  &  \textbf{1.94}   & 3.35  & 2.71 &  23.95 & 23.72   & 2.81  & 1.93\\
\midrule
Precipitation &  \textbf{71.59}  &  \textbf{}   & 70.48  &  &  96.07 &   & 100.00  & \\
\midrule
Flu &  \textbf{19.43}  &  \textbf{}   & 22.59  &  &  19.01 &    & 15.54  & \\
\midrule
Dengue (K=5) &  \textbf{34.67}  &  \textbf{}   & 37.23  & &  33.91 &    & 35.92  & \\
\midrule
Apnea  &  \textbf{27.25}  &  \textbf{}   & 31.47 &  &  30.10 &    & 30.08  & \\
\bottomrule
\end{tabular}

\end{tabular}
\end{center}
}
\end{table*}
}

\newcolumntype{L}[1]{>{\raggedright\let\newline\\\arraybackslash\hspace{0pt}}m{#1}}
\newcolumntype{C}[1]{>{\centering\let\newline\\\arraybackslash\hspace{0pt}}m{#1}}
\newcolumntype{R}[1]{>{\raggedleft\let\newline\\\arraybackslash\hspace{0pt}}m{#1}}

\newcolumntype{?}{!{\vrule width 1pt}}
\newcommand\Tstrut{\rule{0pt}{3ex}}
\newcommand{\ErrTable}{
\begin{table*}[!t]
\caption{{\small Comparison of short- and long-term prediction error (NRMSE\%) on the test sets of real data.}}
\centering
{\scriptsize
\setlength\tabcolsep{1pt}
\begin{tabular}{l?cc|cc|cc|c|c|c|c|c?c|c|c|c|c}
\toprule
&\multicolumn{11}{c?}{{\small Short-term}}
&\multicolumn{5}{c}{{\small Long-term}}
\\
\cline{2-17}
\Tstrut
\multirow{2}{*}{\backslashbox{Dataset}{Model}} 
&\multicolumn{2}{c|}{DSARF} 
&\multicolumn{2}{c|}{rSLDS} 
&\multicolumn{2}{c|}{SLDS} 
& BTMF 
& B-TRMF 
& TRMF 
& RKN
& LSTNet
&DSARF
&BTMF
&B-TRMF
&TRMF
&LSTNet
\\
& \tiny{w/ switch}   & \tiny{w/o switch}& \tiny{w/ switch}   & \tiny{w/o switch}   & \tiny{w/ switch}   & \tiny{w/o switch}  &   & & &&&&&&&\\
\midrule
Birmingham (K=10)\;\;  
&\textbf{5.70} 
&\textbf{\;\;5.76$^*$} 
& 14.23 
& 14.69 
& 8.69
& 14.52 
& 15.27
& 15.84 
& 17.13
& 11.92
& 9.32
&  \textbf{15.05} & 18.02&  28.71 & 22.65 & 23.38\\
Guangzhou (K=30) &  10.21 & 10.20 & \textbf{\;\;10.10$^*$} & 10.11 & 10.86 & 10.16 & 10.30 &  10.75 & 10.83& 10.33
& \textbf{9.17} &
13.01 & \textbf{12.83}&  16.03 & 14.75 & 15.76\\
Hangzhou (K=10) & 17.31 & \textbf{15.55}  &  16.57 & 17.27 & 17.20 & 17.27 & 16.67 &  18.26 & 17.86& \textbf{\;\;16.39$^*$} 
& 16.40 &  
\textbf{15.64} & 18.33 &  20.92 & 17.85 & 16.68\\
Seattle (K=30) &  7.60 & \textbf{7.52}  &  7.54 & \textbf{7.52} & 8.44 & 7.53  & 7.69 &  8.10 & 8.30& 7.61
& 7.73 &
\textbf{14.14}& 14.33 &  22.51 & 16.79 & 16.30\\
PST (K=50) &  1.96 & 2.12  & 1.80 & \textbf{1.75} & 1.94 & \textbf{\;\;1.74$^*$}  & 3.35 &  23.95 & 2.81& 2.25 
& 2.16 &
\textbf{2.53} & 7.43 &  6.91 & 3.49 & 3.17\\
Flu (K=10) &  \textbf{\;\;16.51$^*$} & 16.77  &  18.98 & 18.41 &19.50 &18.25   & 22.59 &  19.01 & \textbf{15.54}& 24.03
& 17.78 &
\textbf{34.96} & 94.31  & 51.12 & 40.87 & 42.11 \\
Dengue (K=5) &  35.29 & \textbf{\;\;34.67$^*$}  &  43.78 & 41.48& 40.92 & 39.91   & 37.23 & \textbf{33.91} & 35.92 & 37.02
& 36.39 &
\textbf{52.83} & 63.85  & 61.04 & 57.34 & 60.46\\
Bat (K=5) & \textbf{7.74} & \textbf{\;\;8.08$^*$}  &  9.91 & 10.11 & 11.13 & 10.19  & 8.89 &  8.16 & 9.02 & 18.59 
& 16.55 &
\textendash&\textendash&\textendash&\textendash&\textendash\\
Precipitation (K=20)&  \textbf{67.41} & 69.52  &  \textbf{\;\;67.70$^*$} & 67.70 & 68.62& 68.44 & 70.48 &  96.07 & 98.01 & 78.80
& 74.35 &
\textendash&\textendash&\textendash&\textendash&\textendash\\
Apnea  &  \textbf{23.86} & \textendash &  27.35 & \textendash & 28.06 & \textendash & 31.47 &  30.10 & 30.08 & \textbf{\;\;27.13$^*$} 
& 27.23 &
\textendash&\textendash&\textendash&\textendash&\textendash\\

\bottomrule
\multicolumn{10}{l}{\tiny The two best results are highlighted in bold fonts without and with asterisk, respectively.}
\end{tabular}
}
\label{tbl:ErrTable}
\end{table*}
}

\newcommand{\DataTable}{
\begin{table}[!b]
\vspace{-7pt}
\caption{{\small Description of real data (see \suppref{data_description} for details).}}
\centering
{\scriptsize
\setlength\tabcolsep{1pt}
\begin{tabular}{l?c|c|c}
\toprule
\Tstrut
Dataset (missing\%) 
& Resolution$^1$
& NT$\times$D
& ${\text{T}_\text{test}}^2$
\\
\midrule
Birmingham Parking (14.89\%) {\tiny\citep{B}} 
&\;q30min for 77d\;
&1386$\times$30
&126 (7d)
\\
Guangzhou Traffic (1.29\%) {\tiny\citep{UTSD}}
& q10min for 61d
& 8784$\times$214 &
720 (5d)
\\
Hangzhou Metro {\tiny\citep{HIPF}} 
& q10min for 25d
&2700$\times$80 
& 540 (5d)
\\
Seattle Traffic {\tiny\citep{SILD}}
& q5min for 28d
& 8064$\times$323 
& 1440 (5d)
\\
Pacific Surface Temp. (PST) {\tiny\citep{IRI/LDEO}}
& qmt for 33yr
& 396$\times$(30$\times$84) 
& 60 (5yr)
\\
Google Flu (9.53\%) {\tiny\citep{Flu}}
& qwk for 13yr
& 658$\times$29 &
84 (2yr)
\\
Google Dengue (4.89\%) {\tiny\citep{Dengue}}
& qwk for 13yr
& 658$\times$10 
& 84 (2yr)
\\
Bat Flight (32.55\%) {\tiny\citep{bergou2015falling}}
& q33msec
& 3303$\times$(34$\times$3) &
700
\\
Precipitation (50.65\%) {\tiny\citep{NOAA}}
& qd for 5yr
& 1462$\times$239
& 305 (1yr)
\\
Apnea {\tiny\citep{rigney1994multichannel,goldberger2000physiobank}}
& q500msec
& 2000$\times$1 
& 1000
\\
\bottomrule
\multicolumn{4}{l}{\tiny $^1$q:every, d:days, yr:years, mt:months, wk:weeks. $^2$The last $\text{T}_\text{test}$ time points are held out for test.}
\end{tabular}
}
\label{tbl:DataTable}
\end{table}
}

\newcommand{\SpatialTable}{
\begin{table}[!t]
\caption{{\small Comparison of test set log-likelihood of spatial factors for real-world data.}}
\centering
{\scriptsize
\setlength\tabcolsep{2pt}
\begin{tabular}{l?c|c|c}
\toprule
\Tstrut
\backslashbox{Dataset}{Spatial Prior} 
& DSARF Hierarchical Prior
& Multivariate Normal Prior
\\
\midrule
Birmingham
& $\mathbf{-1.55\times10^{-1}}$
& $-3.37\times10^{-1}$
\\
Guangzhou
& $\mathbf{-1.74}$
& $-1.96$
\\
Hangzhou
& $\mathbf{-1.95}$
& $-2.45$
\\
Seattle
& $\mathbf{-2.00}$
& $-2.49$
\\
PST
& $\mathbf{-3.99\times10^{-1}}$
& $-1.15$
\\
Flu
& $\mathbf{-5.82\times10^{-2}}$
& $-1.35\times10^{-1}$
\\
Dengue
& $\mathbf{-2.27}$
& $-2.76$
\\
Bat
& $\mathbf{-2.39\times10^{-1}}$
& $-3.50\times10^{-1}$
\\
Precipitation
& $\mathbf{-4.07\times10^{-1}}$
& $-7.56\times10^{-1}$
\\
\bottomrule
{\footnotesize\textbf{Average}}
& $\mathbf{-1.02}$
& $-1.37$
\\
\bottomrule
\multicolumn{3}{l}{\tiny The values are normalized by K$\times$D (i.e., the unit is nats/spatial element).}
\end{tabular}
}
\label{tbl:spatialTable}
\end{table}
}
\newcommand{\supp}{
\newpage
\onecolumn
\newcommand{\beginsupplement}{%
        \setcounter{table}{0}
        \setcounter{equation}{0}
        \renewcommand{\theequation}{S\arabic{equation}}
        \setcounter{section}{0}
        \renewcommand{\thetable}{S\arabic{table}}%
        \setcounter{figure}{0}
        \renewcommand{\thefigure}{S\arabic{figure}}%
}
\newcommand\independent{\protect\mathpalette{\protect\independenT}{\perp}}
\def\independenT##1##2{\mathrel{\rlap{$##1##2$}\mkern2mu{##1##2}}}
\renewcommand\thesection{\Alph{section}}
\beginsupplement
\section{Supplementary Materials}
We have submitted the source code and experiments/datasets along with our submission as part of its supplementary materials. In the following sections, we provide the details on ELBO derivation, synthetic experiments, and hyper-parameter setup for real-world datasets.
\subsection{Derivation of Evidence Lower BOund (ELBO) for DSARF}
\label{app:derivation}
We derive the lower bound on the log-likelihood of observations, $\mathcal{L}(\theta,\phi)$, by incorporating conditional independencies inferred from the graphical model of DSARF in \figref{graph} as follows:
{\small
\begin{align*}
    X_n\independent X_{\neg n}&|w_{n,1:T},F \\
    w_{n,t}\independent w_{ n,\neg(t,t-\ell)},s_{n,\neg t}&|w_{n,t-\ell}, s_{n,t}\\
    w_{n,t}\independent w_{ \neg n},s_{\neg n}&|w_{n,t-\ell}, s_{n,t}\\
    s_{n,t}\independent s_{n,\neg(t,t-1)}&|s_{n,t-1}\\
    s_{n,t}\independent s_{\neg n}&|s_{n,t-1}
\end{align*}
}Considering these conditional independencies, the joint distribution of observations and latent variables will be: 
{\small
\begin{align}
     p_\theta(X,\mathcal{S},W,z,F) = p_\theta(F|z)p_\theta(z)&\prod_{n=1}^N p_\theta(X_n|w_{n,1:T}, F) p_\theta(w_{n,-\ell}) p_\theta(s_{n,0})\nonumber\\
     &\prod_{t=1}^T p_\theta(s_{n,t}|s_{n,t-1}) p_\theta(w_{n,t}|w_{n,t-\ell},s_{n,t})
     \label{eqn:gen}
\end{align}
}We assume a fully factorized variational distribution on $\{\mathcal{S},W,z,F\}$, hence:
{\small
\begin{align}
     q_\phi(\mathcal{S},W,z,F) = q_\phi(F)q_\phi(z)
     \prod_{n=1}^N q_\phi(w_{n,-\ell}) q_\phi(s_{n,0}) \prod_{t=1}^T
     q_\phi(s_{n,t}) q_\phi(w_{n,t})\label{eqn:var}
\end{align}
}We then derive the ELBO by writing down the log-likelihood of observations, and plugging in $p_\theta(\cdot)$ and $q_\phi(\cdot)$ from \eqnref{gen} and \eqnref{var} respectively (we denote continuous latent variables collectively as $\mathcal{Z}=\{W,z,F\}$ for brevity):
{\small
\begin{align}
    \log p_\theta(X) &= \log \sum_\mathcal{S}\int\displaylimits_{\mathcal{Z}}q_\phi(\mathcal{S},\mathcal{Z}) \frac{p_\theta(X,\mathcal{S},\mathcal{Z})}{q_\phi(\mathcal{S},\mathcal{Z})} d\mathcal{Z}
    \underset{\textbf{\{Jensen's inequality\}}}{\geq}
    \sum_\mathcal{S}\int\displaylimits_{\mathcal{Z}}q_\phi(\mathcal{S},\mathcal{Z}) \log\frac{p_\theta(X,\mathcal{S},\mathcal{Z})}{q_\phi(\mathcal{S},\mathcal{Z})} d\mathcal{Z}\nonumber\\
    &=\sum_\mathcal{S}\int\displaylimits_{\mathcal{Z}} q(\mathcal{S},\mathcal{Z})
    \log\frac{p(F|z)p(z)}{q(F)q(z)}\prod\limits_{n=1}^N\frac{p(X_n|w_{n,1:T}, F)  p(w_{n,-\ell})p(s_{n,0})}{q(w_{n,-\ell})q(s_{n,0})}\nonumber\\
    &\quad\quad\quad\quad\quad\quad\quad\quad\quad\quad\quad\quad\quad\quad\;
    \prod\limits_{t=1}^T\frac{p(w_{n,t}|w_{n,t-\ell},s_{n,t}) p(s_{n,t}|s_{n,t-1})}{ q(w_{n,t})q(s_{n,t})} d\mathcal{Z}\nonumber\\
    &=\sum_{\mathcal{S}\backslash s^\prime} q(\mathcal{S}\backslash s^\prime)\int\displaylimits_{\mathcal{Z}\backslash z^\prime}q(\mathcal{Z}\backslash z^\prime) d\mathcal{Z}\backslash z^\prime\,\boldsymbol{\Bigg(}\,
    \int\displaylimits_{\underset{s^\prime=\varnothing}{z^\prime=\{z, F\}}}q(z)q(F)\log\frac{p(F|z)}{q(F)} dz^\prime+\int\displaylimits_{\underset{s^\prime=\varnothing}{z^\prime=\{z\}}}q(z)\log\frac{p(z)}{q(z)} dz^\prime\nonumber\\
    &\quad
    +\mathlarger{\boldsymbol{\sum_{n=1}^{N}}}\,
    \int\displaylimits_{\underset{s^\prime=\varnothing}{z^\prime=\{w_{n,1:T},F\}}} q(w_{n,1:T},F)\log p(X_n|w_{n,1:T},F) dz^\prime\nonumber\\
    &\quad
    \quad\quad\quad
    +\sum_{\underset{z^\prime=\varnothing}{s^\prime=\{s_{n,0}\}}}q(s_{n,0})\log \frac{p(s_{n,0})}{q(s_{n,0})}
    +\int\displaylimits_{\underset{s^\prime=\varnothing}{z^\prime=\{w_{n,-\ell}\}}}q(w_{n,-\ell})\log\frac{p(w_{n,-\ell})}{q(w_{n,-\ell})} dz^\prime\nonumber\\
    &\quad
    +
    \mathlarger{\boldsymbol{\sum_{t=1}^T}} 
    \sum_{\underset{z^\prime=\varnothing}{s^\prime=\{s_{n,t}\,,\,s_{n,t-1}\}}}q(s_{n,t-1})q(s_{n,t})\log\frac{p(s_{n,t}|s_{n,t-1})}{q(s_{n,t})}\nonumber\\
    &\quad
    \quad\quad\quad
    +\sum_{s^\prime=\{s_{n,t}\}}
    \int\displaylimits_{z^\prime=\{w_{n,t-\ell}\}}q(w_{n,t-\ell})\log\frac{p(w_{n,t}|w_{n,t-\ell},s_{n,t})}{q(w_{n,t})} dz^\prime\boldsymbol{\Bigg)},
    \label{eqn:sup1}
\end{align}
}where the integrations/summations over distributions of the latent variables outside of the big parenthesis is: $\sum_{\mathcal{S}\backslash s^\prime} q(\mathcal{S}\backslash s^\prime)\int_{\mathcal{Z}\backslash z^\prime}q(\mathcal{Z}\backslash z^\prime)d\mathcal{Z}\backslash z^\prime = 1$ for $\{s^\prime, z^\prime\}$ of each term inside the parenthesis (we are abusing the notation here for the sake of brevity). Considering that $\mathbb{KL}(q,p) = \int q\log \frac{q}{p}$, and $\mathbb{E}_{q(x)}[f(x)] = \int_x q(x)f(x)$, we rewrite each term of the summations in \eqnref{sup1} to summarize the ELBO:
{\small
\begin{align}
 \mathcal{L}(\theta, \phi)&=\sum_{n=1}^N \Big(\boldsymbol{\mathcal{L}_n^\textbf{rec}}+ \boldsymbol{\mathcal{L}_n^\text{$s_0,w_{-\ell}$}}+\sum_{t=1}^T\big(\boldsymbol{\mathcal{L}_{t,n}^{\mathcal{S}}}+\boldsymbol{\mathcal{L}_{t,n}^{W}}\big)\Big)+\boldsymbol{\mathcal{L}^\textbf{F}}\label{eqn:sup2},\\
 \boldsymbol{\mathcal{L}_n^\textbf{rec}} =& \mathbb{E}_{q(w_{n,1:T},F)}\Big[\log p(X_n|w_{n,1:T},F)\Big]\nonumber\\
 \boldsymbol{\mathcal{L}_n^\text{$s_0,w_{-\ell}$}} =& -\text{KL}\big(q(s_{n,0})||p(s_0)\big)
 -\text{KL}\big(q(w_{n,-\ell})||p(w_{-\ell})\big)\nonumber\\
 \boldsymbol{\mathcal{L}_{n,t}^{S}} =& -\mathbb{E}_{q(s_{n,t-1})}\Big[\text{KL}\big(q(s_{n,t})||p(s_{n,t}|s_{n,t-1})\big)\Big]\nonumber\\
 \boldsymbol{\mathcal{L}_{n,t}^{\mathcal{W}}} =& -\sum_{s_{n,t}} q(s_{n,t})\mathbb{E}_{q(w_{n,t-\ell})} \Big[\text{KL}\big(q(w_{n,t})\|p\big(w_{n,t}|w_{n,t-\ell},s_{n,t})\big)\Big]\nonumber\\
 \boldsymbol{\mathcal{L}^F}=& -\mathbb{E}_{q(z)}\Big[\text{KL}\big(q(F)||p(F|z)\big)\Big]
 -\text{KL}\big(q(z)||p(z)\big).\nonumber
\end{align}
}\textbf{Kullback Leibler ($\text{KL}$) Divergence Terms:}
We can analytically calculate the $\text{KL}$ terms of ELBO in \eqnref{sup2}. For two multivariate ($d$-dimensional) Gaussian distributions $q(\cdot)$, and $p(\cdot)$, the $\text{KL}$ divergence is:
{\small
\begin{align*}
   \text{KL}(q, p) = \frac{1}{2} \Big[&\log\frac{|\Sigma_p|}{|\Sigma_q|} - d + \text{tr}(\Sigma_p^{-1} \Sigma_q)
   + (\mu_p - \mu_q)^T \Sigma_p^{-1} (\mu_p-\mu_q)\Big],
\end{align*}
}which would be further simplified here, in the case of diagonal covariances. The $\text{KL}$ divergence between two $S$-dimensional categorical distributions, $q(\cdot)$, and $p(\cdot)$, can be readily computed:
\begin{align*}
    \text{KL}(q,p) = \sum_s q_s \log\Big(\frac{q_s}{p_s}\Big),
\end{align*}
where the subscript $s$ denotes the probability of state $s$ out of $S$ possible states.

\subsection{Normalized root-mean-square error (NRMSE\%)}
\label{app:nrmse}
We compute the prediction NRMSE\% for a test set $X_{T\times D}$ as follows:
\begin{align*}
    \text{NRMSE\%} = \frac{\sqrt{\frac{1}{T.D}\sum_{t=1}^{T}\sum_{d=1}^D (X_{t,d}-\hat{X}_{t,d})^2}}{\sigma_X}\times 100
\end{align*}
where $\hat{X}$ are the predicted values, and $\sigma_X$ is the global standard deviation of the test set matrix, $X$. 
\subsection{More details on the synthetic experiments}
\label{app:synthetic_details}
For all the synthetic experiments in \secref{synthetic_exp} (toy example, Lorenz attractor and double pendulum), we observed the continuous latents, $W$, through a linear projection and an additive noise as: $X=W^\top F+\mathcal{N}(0,\sigma\,\mathbf{I})$. Fitting DSARF (and the other baselines) on these observations, $X$, would result in a projected estimation of $W$. While we report the prediction NRMSE\% in the observation space, for the sake of a more informative visualization in \figref{synthetic}, we project the actual $W$ to the estimated space of the model as follows:
\begin{align*}
    &X = W^\top F,\quad \text{where $F$ are the actual spatial factors}\\
    &X = W_{\textbf{proj}}^\top \hat{F},\quad \text{where $\hat{F}$ are the inferred spatial factors}\\
    &\text{Then,}\\
    &W^\top F = W_{\textbf{proj}}^\top \hat{F}
    \xrightarrow{\text{Least-squares solution}} W_{\textbf{proj}} = (F\hat{F}^{+})^\top W
\end{align*}
where $\hat{F}^{+}$ is the Moore–Penrose inverse of $\hat{F}$, the inferred spatial factors after fitting the model. We visualized $W_{\textbf{proj}}$ as groundtruth along with the inferred temporal latents, $\hat{W}$, in \figref{synthetic}.
\\
\\
\textbf{Double pendulum:}
We have additionally visualized the inferred dynamical trajectory in $\theta_1$--$\theta_2$ space for $S=3$ 
in \figref{penfigsupp} which shows a rough segmentation along $\theta_1$, the deflection angle of the first pendulum. Note that we had visualized this inferred dynamical trajectory in Cartesian space for $S=3$ in \figref{synthetic}c in the paper.    
\penfigsupp
\subsection{Description of real-world datasets}
\label{app:data_description}
\textbf{Sleep Apnea}, A physiological data set from a patient tentatively diagnosed with sleep apnea \citep{rigney1994multichannel,goldberger2000physiobank}. The respiration pattern in sleep apnea is characterized by at least two regimes: no breathing and gasping breathing induced by a reflex arousal. We used samples 6201-7200 for training and 5201-6200 for testing as in \citet{ghahramani1996switching}.\\
\textbf{Bat Flight}, \citep{bergou2015falling}: Registered 3D coordinates of $34$ markers on a Bat body (i.e., joint locations) recorded over time during a landing maneuver for $10$ experimental runs with $32.55\%$ missing values as markers are frequently occluded during the flight. We organized this dataset into a tuple of $(T_n\times 34\times 3)_{n=1}^{10}$ tensors, where $T_n$ ranges from $166-436$ time points, and left the last two runs for test.\\
\textbf{Birmingham Parking}, \citep{B}: Recorded occupancy of 30 car parks in Birmingham, UK, from October 4 to December 19, 2016 (77 days) every half an hour (18 time intervals per day) with 14.89\% missing values. We picked the last 7 days for test.\\
\textbf{Hangzhou Metro}, \citep{HIPF}: Incoming passenger flow from 80 metro stations in Hangzhou, China, from January 1 to January 25, 2019 (25 days) with a 10-minute resolution during service hours (108 time intervals per day). We picked the last 5 days for test.\\
\textbf{Guangzhou Traffic}, \citep{UTSD}: Recorded traffic speed data from 214 road segments in Guangzhou, China, from August 1 to September 30, 2016 (61 days) with a 10-minute resolution (144 time intervals per day) with 1.29\% missing values. We picked the last 5 days for test.\\
\textbf{Seattle Traffic}, \citep{SILD}: Recorded traffic speed from 323 loop detectors in Seattle, USA, from January 1 to January 28, 2015 (28 days) with a 5-minute resolution (288 time intervals per day). We picked the last 5 days for test. \\
\textbf{Pacific Surface Temperature (PST)}, \citep{IRI/LDEO}: Gridded (at a 2 by 2 degrees resolution, corresponding to 2520 spatial locations) monthly sea surface temperature on the Pacific for 396 consecutive months from January 1970 through December 2002. We held the last 5 years for test.\\
\textbf{Colorado Precipitation}, \citep{NOAA}: Daily precipitation measurements for $239$ areas around Boulder Colorado for the period of November 2013– November 2017 with $50.65\%$ missing values. We held the last year for test.\\
\textbf{Google Flu}, \citep{Flu}: Weekly estimates of influenza activity in $29$ countries for 13 years from 2003-2016 with $9.53\%$ missing values. We kept the last 2 years for test.\\
\textbf{Google Dengue}, \citep{Dengue}: Weekly estimates of Dengue trend in $10$ countries for 13 years from 2003-2016 with $4.89\%$ missing values. We kept the last 2 years for test.

\subsection{Hyperparameter setup for the real-world datasets}
\label{app:hyperparameters}
We set the number of factors $K=\{10, 30, 10, 30, 50, 10, 5, 5, 20, 1\}$ and the number of discrete states $S=\{3, 3, 3, 3, 2, 2, 2, 3, 2, 2\}$ for Birmingham, Guangzhou, Hangzhou, Seattle, PST, Google flu, Google Dengue, Bat flight, Precipitation and Apnea datasets respectively for DSARF and the other baselines accordingly (if applicable). For rSLDS, SLDS, and RKN we set the dimension of the continuous latent space to $K$.
The choice of $K$ for Birmingham, Guangzhou, Hangzhou and Seattle datasets is consistent with \citet{sun2019bayesian}. For other datasets the choice of $K$ is well reasoned, and is determined with probabilistic principal component analysis (PPCA). For example, in the Bat flight dataset we choose $K=5$ as the movement of joints in bat are highly correlated during flight as suggested in \citet{riskin2008quantifying}, where they found that approximating the bat’s motion with only one
third of the principal components accounted for 95\% of the
variance of the original behavior. We performed a manual search for other hyperparameters ($S$, and $\ell$) based on apriori knowledge about a dataset.

We used lag set $\ell=\{1,2\}$ for DSARF on all short-term prediction experiments on real-world data (set accordingly for BTMF, TRMF, B-TRMF, and LSTNet). For long-term predictions, we used lag set $\ell=\{1, 2, 3, T_0, T_0+1, T_0+2, 7T_0, 7T_0+1, 7T_0+2\}$ for Birmingham, Guangzhou, Hangzhou and Seattle datasets as in \citet{sun2019bayesian} (where $T_0$ is the time points per day), $\ell=\{1,2, 52, 52+1, 2\times52, 2\times52+1\}$ weeks for Google flu and Dengue datasets and $\ell=\{1, 2, 12, 12+1, 6\times12, 6\times12+1\}$ months for the PST dataset.

\subsection{Spatial generative model}
\label{app:spatial_generative}
In order to justify the effect of our proposed spatial generative modeling, we compared that with the widely used multivariate Normal prior for spatial factors. To this end, we fit DSARF once with the proposed hierarchical prior, and another time with a multivariate Normal prior on a train set, and then computed an estimation of the expected log-likelihood (LL) of the spatial factors on a test set as follows:
{\small

\begin{align*}
  \text{LL} &= \mathbb{E}_{F\,\sim\,q(F)}\left[\log\,\mathcal{N}(F\,;\mu_\theta^F, \Sigma_\theta^F)\right],
  \quad\text{for a multivariate Normal prior}\\
  \text{LL} &= \mathbb{E}_{F\,\sim\,q(F)}\mathbb{E}_{z\,\sim\,q(z)}
  \left[\log\,\mathcal{N}\left(F\,;\mu_\theta^F(z), \Sigma_\theta^F(z)\right)\right],
  \quad\text{for the proposed hierarchical prior}
\end{align*}
}where we estimated each expectation with $100$ samples. The results of this comparison on our real-world data are summarized in \tblref{spatialTable}, and demonstrate that DSARF resulted in spatial factors with higher test set log-likelihood (i.e, higher posterior predictive probability) on all of our real-data experiments, when compared to the multivariate Normal prior.
\SpatialTable
}

\title{Deep Switching Auto-Regressive Factorization:\\ Application to Time Series Forecasting}

\begin{document}

\maketitle
\begin{abstract}
We introduce deep switching auto-regressive factorization (DSARF), a deep generative model for spatio-temporal data with the capability to unravel recurring patterns in the data and perform robust short- and long-term predictions. Similar to other factor analysis methods, DSARF approximates high dimensional data by a product between time dependent weights and spatially dependent factors. These weights and factors are in turn represented in terms of lower dimensional latent variables that are inferred using stochastic variational inference. DSARF is different from the state-of-the-art techniques in that it parameterizes the weights in terms of a deep switching vector auto-regressive likelihood governed with a Markovian prior, which is able to capture the non-linear inter-dependencies among weights to characterize multimodal temporal dynamics. This results in a flexible hierarchical deep generative factor analysis model that can be extended to (i) provide a collection of potentially interpretable states abstracted from the process dynamics, and (ii) perform short- and long-term vector time series prediction in a complex multi-relational setting. Our extensive experiments, which include simulated data and real data from a wide range of applications such as climate change, weather forecasting, traffic, infectious disease spread and nonlinear physical systems attest the superior performance of DSARF in terms of long- and short-term prediction error, when compared with the state-of-the-art methods\footnote{The source code and experiments are provided as part of the supplementary submission.}.
\end{abstract}

\section{Introduction}
\label{sec:intro}
Ever-improving sensing technologies offer fast and accurate collection of large-scale spatio-temporal data in various applications, ranging from medicine and biology to marketing and traffic control. In these domains, modeling the temporal dynamics and spatial relations of data have been investigated and analysed from different perspectives. \par
As these multivariate spatio-temporal data often exhibit high levels of correlation between dimensions, they can naturally be thought of as governed by a smaller number of underlying components. Tensor/matrix factorization frameworks are used to describe variability in these correlated dimensions in terms of potentially lower dimensional unobserved variables, namely temporal \emph{weights} and spatial \emph{factors}. Accordingly, Bayesian probabilistic global matrix/tensor factorization has been investigated in \citet{salakhutdinov2008bayesian,zhao2015bayesian,chen2019missing} for time series imputation. Besides, linear temporal dynamics have been adapted into this framework in \citet{xiong2010temporal,charlin2015dynamic,sun2019bayesian}. A number of non-Bayesian dynamical matrix factorization methods have been explored in \citet{rogers2013multilinear,sun2014collaborative,bahadori2014fast,cai2015facets,yu2016temporal,takeuchi2017autoregressive,jing2018high}. Amongst these methods, some assume a linear vector auto-regressive model for the temporal weights \citep{bahadori2014fast, yu2016temporal}, and spatial factors \citep{takeuchi2017autoregressive} to model higher-order auto-regressive dependencies in multivariate time series data.

From another perspective, Bayesian switching linear state-space models, \citep{chang1978state, hamilton1990analysis, ghahramani1996switching, murphy1998switching, fox2009nonparametric, linderman2017bayesian, nassar2019tree,becker2019switching}, have provided a more flexible structure for modeling temporal dynamics characterized by several modes. These models are specifically useful in the applications where complex dynamical behaviors can be broken down into simpler potentially interpretable units, which in turn provides additional insight into the rich processes generating complex natural phenomena. These models achieve globally nonlinear dynamics by composing linear systems through switching, \citep{sontag1981nonlinear}. Besides, Gaussian state space models adapting neural networks have been used for approximating nonlinear first-order temporal dynamics in  \citet{krishnan2015deep,watter2015embed,karl2017deep,krishnan2017structured,fraccaro2017disentangled,becker2019recurrent,farnoosh2020deep}. 

In this paper, we introduce deep switching auto-regressive factorization (DSARF) in a Bayesian framework. Our method adds to the current body of knowledge by extending switching linear dynamical system models and Bayesian dynamical matrix factorization methods, and combining their favorable properties. Specifically, for temporal dynamic modeling, we employ a non-linear vector auto-regressive latent model parameterized by neural networks and governed by a Markovian chain of discrete switches to capture higher-order multimodal latent dependencies. This will provide a more flexible model that expands prediction horizon and improves long- and short-term forecasting. In addition, we use a deep generative prior for estimation of multimodal distributions for spatial factors. We leverage the tensor/matrix factorization framework to make our model scalable to high dimensional data and solve this model efficiently with approximate variational inference.\par
Our hierarchical generative model with the help of corresponding learning and inference algorithm is able to handle missing data and to provide uncertainty measures for estimations. We demonstrate our model performance using experiments including simulated and real-world data from a wide range of application areas. Our experiments show that DSARF achieves better predictive performance on unseen data relative to current state-of-the-art methods when evaluated based on short- and long-term prediction errors.
In the following section, we provide more motivation and background regarding our main contributions. 

\section{Background and Motivation}
\label{sec:Backg}
Our hierarchical model consists of three components: switching dynamical systems, non-linear vector auto-regression, and tensor/matrix factorization, for which we provide reviews and justifications in this section. \par
\textbf{Linear Gaussian dynamical systems} operating in Markov dependent switching environment have long been investigated in the literature, \citep{ackerson1970state, chang1978state, hamilton1990analysis, ghahramani1996switching, murphy1998switching, fox2009nonparametric}. These models, also known as switching linear dynamical system (SLDS), decompose nonlinear time series data into series of simpler, repeated dynamical modes. The SLDS model learns the underlying nonlinear generative process of the data as it breaks down the data sequences into coherent, potentially interpretable, discrete units, similar to the \emph{piecewise affine} (PWA) framework in control systems \citep{sontag1981nonlinear, juloski2005bayesian, paoletti2007identification} . The generative process starts with sampling a discrete latent state $s_t \in \{1, \dots, S\}$ at each time $t = 1, \dots, T$ according to Markovian dynamics $s_t\mid s_{t-1},\mathbf{\Phi} \sim \; \pi_{s_{t-1}}$, where $\mathbf{\Phi}$ is the Markov transition matrix and $\pi_s$ is the categorical distribution parameter. Then, a continuous latent state $w_t \in \mathbb{R}^K$ is sampled  from a normal distribution whose mean follows a conditionally linear dynamics as $w_t = \mathbf{A}_{s_t} w_{t-1}+\mathbf{b}_{s_t}+\nu_{t-1}, \; \nu_{t-1} \overset{iid}{\sim} \mathcal{N}(0,\mathbf{Q}_{s_t})$, for matrices $\mathbf{A}_s, \mathbf{Q}_s \in \mathbb{R}^{K\times K}$ and vectors $\mathbf{b}_s \in \mathbb{R}^K$ for $s = 1, 2, \dots, S$. Finally, a linear Gaussian observation $x_t \in \mathbb{R}^D$ is generated from the continuous latent state $w_t$ according to $x_t = \mathbf{C}_{s_t} w_{t}+\mathbf{d}_{s_t}+\mu_{t}, \; \mu_{t} \overset{iid}{\sim} \mathcal{N}(0,\mathbf{G}_{s_t})$, for matrices $\mathbf{C}_s \in \mathbb{R}^{D\times K}, \mathbf{G}_s \in \mathbb{R}^{D\times D}$ and vectors $\mathbf{d}_s \in \mathbb{R}^D$. SLDS parameters are learned in a Bayesian inference approach. In this framework, the probabilistic dependencies are in such a way that $s_{t+1}\mid s_t$ is independent of the continuous state $w_t$, and hence the model cannot learn the transition of the discrete latent state when continuous latent state enters a particular region of state space. This problem is addressed in recurrent switching linear dynamical system (rSLDS), \citep{linderman2017bayesian,nassar2019tree} by allowing the discrete state transition probabilities to depend on the preceding continuous latent
state, i.e, $s_t|s_{t-1},w_{t-1}$. 
rSLDS studies proposed to use auxiliary variable methods for approximate inference in a multi-stage training process.
\citet{nassar2019tree} extended rSLDS of \citet{linderman2017bayesian} by enforcing a tree-structured prior on the switching variables in which subtrees share similar dynamics. \citet{becker2019switching} proposed to learn an rSLDS model through a recurrent variational autoencoder (rVAE) framework, and approximated switching variables by a continuous relaxation. This amortized inference compromised the applicability of their model on missing data, as they only included physics-simulated experiments.
\par
The rSLDS prediction horizon is, however, limited as it adopts the first-order linear Markovian dynamics, a prevalent model in the literature, for both discrete and continuous latent state, $s_t$ and $w_t$. On the other hand, we advocate the use of higher-order dependencies in an auto-regressive approach, as in \citet{bahadori2014fast, sun2019bayesian}, namely we introduce  deep generative \textbf{vector auto-regressive} priors for the continuous latent variable, $w_t$, which gives more flexibility to the model to expand its prediction horizon and capture higher-order nonlinear auto-regressive relations amongst its continuous latent variable. More specifically, we use a weighted combination of a linear and a non-linear transformation as the relation between two continuous latent variables $w_t$ and $w_{t-\ell}$, where $\ell$ is a lag set. \par
\textbf{Bayesian tensor/matrix factorization} constitutes the outermost layer of our hierarchical probabilistic model offering an effective approach to convert massive data into a lower-dimensional and computationally more tractable set of latent, e.g., temporal and spatial components. Tensor/matrix factorization frameworks, \citep{sun2014collaborative,bahadori2014fast,cai2015facets,zhao2015bayesian,yu2016temporal,takeuchi2017autoregressive,chen2019missing}, are used to describe variability in high dimensional correlated data in terms of potentially lower dimensional unobserved variables called \emph{factors}. In other words, given an observation matrix $X\in \mathbb{R}^{T\times D}$ of spatio-temporal data with $T$ time points and $D$ spatial locations, these methods decompose $X$ into a set of $K\ll D$ temporal factors (weights) $W\in \mathbb{R}^{K\times T}$, and spatial factors $F\in \mathbb{R}^{K\times D}$ as $X \approx W^\top F$, where temporal dynamics are modeled in $W$ as first-, e.g., in \citet{sun2014collaborative,cai2015facets}, or higher-order linear dependencies, e.g., in \citet{bahadori2014fast,yu2016temporal,takeuchi2017autoregressive}.
\graphicalmodel

While the focus of this paper is on Bayesian \emph{switching} dynamical modeling, several studies have employed neural networks for non-linear \textbf{Markovian state-space modeling}, \citep{krishnan2015deep,watter2015embed,karl2017deep,krishnan2017structured,fraccaro2017disentangled,becker2019recurrent} (a.k.a. SSMs), and \textbf{multidimansional times series forecasting}, \citep{chang2018memory,lai2018modeling,rangapuram2018deep,li2019enhancing,sen2019think,salinas2020deepar} (denoted by fNNs here). Deep SSMs operate in an encoding/decoding framework (similar to VAEs), and are restricted to first-order Markovian dependencies. fNNs estimate model parameters from input data using recurrent neural networks (RNNs), e.g., in \citet{chang2018memory,lai2018modeling,salinas2020deepar}, Transformers, in \citet{li2019enhancing}, or temporal convolution networks (TCNs), in \citet{sen2019think}. While linear vector auto-regression on high-dimensional input data is adopted in fNNs of \citet{chang2018memory,lai2018modeling}, most fNNs employ first-order autoregressive models. In addition, many SSMs and fNNs are not naturally tractable to data with missing values (without e.g., prior imputation or zero-filling), since, in training, target values, $x_{1:T}$, are fed directly to a neural network for model estimation, e.g, in fNNs, or variational estimation, i.e, amortized inference in SSMs (see \citet{che2018recurrent,ghazi2018robust,rangapuram2018deep,mattei2019miwae} for a discussion). 
As such, the two recent works, \citet{sen2019think,salinas2020deepar}, excluded time series with missing values from their experiments.
This is a major motivation for non-amortized inference in DSARF and some of the comparison baselines in this paper as the datasets in our experiments have up to 50\% missing values. 
In the following section, we provide detailed formulation of DSARF model and its inference procedure. 
\section{Problem Formulation: Deep Switching Auto-Regressive Factorization}
\label{sec:prob}
We consider a corpus of $N$ spatio-temporal data $\{X_n\}_{n=1}^N$, where each $X_n \in \mathbb{R}^{T \times D}$ contains $T$ time points and $D$ spatial locations. We assume that $X_n$ can be decomposed into a weighted summation of $K\ll D$ factors over time:
{\small
\begin{equation}
     X_n \approx [w_1,\cdots, w_T]_n^\top[f_1; \cdots; f_K] = W_n^\top F,
     \label{eqn:fac}
\end{equation}
}where $f_k\in\mathbb{R}^D$ is the $k^{th}$ spatial factor and $w_t\in\mathbb{R}^K$ are weights at time $t$.
In order to model temporal dynamics, we assume that these low dimensional weights, $W=\{w_{t}\}_{t=1}^T$, are generated in accordance with a set of temporal lags, $\ell$, through a deep probabilistic \emph{switching} auto-regressive model, governed by a Markovian chain of \emph{discrete} latent states, $\mathcal{S}=\{s_{t}\}_{t=1}^T$ as follows: $w_t \sim p(w_t|w_{t-\ell},s_t),\; s_t \sim p(s_t|s_{t-1})$.
In other words, in the underlying state-space model of data, $w_t$ is conditioned on $w_{t-\ell}$ (weights at the temporal lags specified in $\ell$), and $s_t$ (state of the model at time $t$). This encourages a \emph{multimodal} distribution for the temporal generative model. We further assume that spatial factors, $F=\{f_{k}\}_{k=1}^K$, come from a shared low dimensional latent variable, $z$, which ensures the estimation of a multimodal distribution for the spatial factors as follows: $f_{1:K}\sim p(F|z),\;z\sim p(z)$.

These assumptions define the graphical model for DSARF in \figref{graph}. We train this hierarchical model using stochastic variational methods \citep{hoffman2013stochastic,ranganath2013adaptive,kingma2013auto,rezende2015variational}. These methods approximate the posterior $p_\theta(\mathcal{S}, W, z, F|X)$ using a variational distribution $q_\phi(\mathcal{S}, W, z, F)$ by maximizing a lower bound (known as ELBO) $\mathcal{L}(\theta,\phi) \le \log p_\theta(X)$:
{\small
\begin{align}
    \mathcal{L}(\theta, \phi)
    &=
    \mathbb{E}_{q_\phi(\mathcal{S}, W, z, F)}
    \left[
    \log 
    \frac{p_\theta(X,\mathcal{S}, W, z, F)}
         {q_\phi(\mathcal{S}, W, z, F)}
    \right]\label{eqn:elbo}
    \\
    &=
    \log p_\theta(X)
    - 
    \text{KL}(q_\phi(\mathcal{S}, W, z, F) \,||\, p_\theta(\mathcal{S}, W, z, F|X)).\nonumber
\end{align}
}By maximizing this bound with respect to the parameters $\theta$, we learn a deep generative model that defines a distribution over datasets $p_\theta(X)$. By maximizing the bound over the parameters $\phi$, we perform Bayesian inference by approximating the distribution $q_\phi(\mathcal{S}, W, z, F) \simeq p_\theta(\mathcal{S}, W, z, F|X)$ over latent variables for each data point.
Considering the proposed generative model, the joint distribution of observations and latents will be
(denoting $\mathcal{Z}=\{W, z, F\}$ for brevity):
{\small
\begin{align}
     \hspace{-.7em}p_\theta(X,\mathcal{S},\mathcal{Z}) = p(F|&z)p(z)\prod_{n=1}^N p(X_n|W_n, F) p(w_{n,-\ell}) p(s_{n,0})\nonumber\\
     &\prod_{t=1}^T p(s_{n,t}|s_{n,t-1}) p(w_{n,t}|w_{n,t-\ell},s_{n,t})
     \label{eqn:pgen}
\end{align}
}We approximate the posterior distributions of latents with a fully factorized variational distribution:
{\small
\begin{flalign}
     \hspace{-.5em}q(\mathcal{S},\mathcal{Z}) = q(F)q(z)
     \prod_{n=1}^N q(w_{n,-\ell}) q(s_{n,0}) \prod_{t=1}^T
     q(s_{n,t}) q(w_{n,t})\hspace{-.2em}
     \label{eqn:qinf}
\end{flalign}
}The ELBO is then derived by plugging in $p_\theta(\cdot)$ and $q_\phi(\cdot)$ from \eqnref{pgen} and \eqnref{qinf} respectively into \eqnref{elbo} (see \suppref{derivation} for derivation details):
{\small
\begin{align}
 &\mathcal{L}(\theta, \phi)=\sum_{n=1}^N \Big(\boldsymbol{\mathcal{L}_n^\textbf{rec}}+ \boldsymbol{\mathcal{L}_n^\text{$s_0,w_{-\ell}$}}+\sum_{t=1}^T\big(\boldsymbol{\mathcal{L}_{t,n}^{\mathcal{S}}}+\boldsymbol{\mathcal{L}_{t,n}^{W}}\big)\Big)+\boldsymbol{\mathcal{L}^\textbf{F}},\nonumber\\
 &\boldsymbol{\mathcal{L}_n^\textbf{rec}} = \mathbb{E}_{q(w_{n,1:T},F)}\Big[\log p(X_n|w_{n,1:T},F)\Big]\nonumber\\
 &\boldsymbol{\mathcal{L}_n^\text{$s_0,w_{-\ell}$}} = -\text{KL}\big(q(s_{n,0})||p(s_0)\big)
 -\text{KL}\big(q(w_{n,-\ell})||p(w_{-\ell})\big)\nonumber\\
 &\boldsymbol{\mathcal{L}_{n,t}^{S}} = -\mathbb{E}_{q(s_{n,t-1})}\Big[\text{KL}\big(q(s_{n,t})||p(s_{n,t}|s_{n,t-1})\big)\Big]\nonumber\\
 &\boldsymbol{\mathcal{L}_{n,t}^{\mathcal{W}}} = -\mathbb{E}_{q(s_{n,t})q(w_{n,t-\ell})} \Big[\text{KL}\big(q(w_{n,t})\|p\big(w_{n,t}|w_{n,t-\ell},s_{n,t})\big)\Big]\nonumber\\
 &\boldsymbol{\mathcal{L}^F}= -\mathbb{E}_{q(z)}\Big[\text{KL}\big(q(F)||p(F|z)\big)\Big]
 -\text{KL}\big(q(z)||p(z)\big).\label{eqn:elbo_terms}
\end{align}
}In the following paragraphs, the parameterization of each term in  \eqnref{elbo_terms} is described.

\textbf{Latent States ($\mathcal{S}$):}
We assume that each data point at a specific time, $x_{n,t}$, belongs to a specific state out of $S$ possible states. This is declared by the categorical variable $s_{n,t}$ in our temporal generative model. These discrete latents, $s_{n,1:T}$, are configured in a Markov chain and govern the state transitions over time as follows ($n$ is dropped hereafter):
{\small
\begin{equation}
    p_\theta(s_{t}|s_{t-1}) = \text{Cat}(\mathbf{\Phi}_\theta\,\mathlarger{\boldsymbol{\pi}}_{s_{t-1}}),\quad q_\phi(s_{t-1}) = \text{Cat}(\mathlarger{\boldsymbol{\pi}}_{s_{t-1}}),\label{eqn:state_transition}
\end{equation}
}where $\mathlarger{\boldsymbol{\pi}}_{s_{t-1}} = [\pi_1, \cdots, \pi_S]$ is the $S$-dimensional posterior parameter vector of $s_{t-1}$, representing probabilities of the categorical distribution, and $\mathbf{\Phi}_\theta\in\mathbb{R}^{S\times S}$ is a valid probability transition matrix.
In practice, we pass $\mathbf{\Phi}_\theta\,\mathlarger{\boldsymbol{\pi}}_{s_{t-1}}$ from a softmax function to ensure a valid probability vector. 

\textbf{Temporal Latents ($W$):}
We adopt a switching Gaussian dynamic for the temporal latent transitions governed by the discrete latent states, $s_t$. In other words, we assume that the marginal distribution of temporal weights, $w_{t}$, follows a Gaussian mixture in the latent space, such that:
{\small
\begin{equation}
   p_\theta(w_{t}|w_{t-\ell},s_{t} = s) = \mathcal{N}\Big(\mu_{\theta_s}^w(w_{t-\ell}), \Sigma_{\theta_s}^w(w_{t-\ell})\Big),\nonumber
\end{equation}
}where $s\in\{1,\cdots,S\}$, and state-specific $\mu_{\theta_s}^w(\cdot)$ and diagonal $\Sigma_{\theta_s}^w(\cdot)$ are parameterized by multilayer perceptrons (MLPs), hence, follow a \emph{nonlinear} vector auto-regressive model given $w_{t-\ell}$, temporal weights in accordance with a lag set $\ell$, as input (e.g., $w_{t-1},w_{t-2}$). Namely, we feed $w_{t-\ell}$ to a multi-head MLP for estimating the Gaussian parameters, e.g.,
{\small
\begin{align}
\mu_{\theta_s}^w=\text{FC}_s(h_s),\quad
h_s = \sum_{l\in\ell}\sigma(\text{FC}_{s,l}(w_{t-l})),\nonumber
\end{align}
}where FC denotes a fully connected layer, and $\sigma$ is a non-linear activation function. We further combine a \emph{linear} vector auto-regression (VAR) of $w_{t-\ell}$ with the estimated mean from neural network to support both linear and nonlinear dynamics:
{\small
\begin{equation}
    \mathlarger{\mu}_{w_t|w_{t-\ell},\,s_{t}=s}  =  (\mathbf{1}-\mathbf{g}_s) \odot \text{VAR}_{\theta_s}(w_{t-\ell})+ \mathbf{g}_s \odot \mu_{\theta_s}^w(w_{t-\ell}),\nonumber
\end{equation}
}where $\odot$ is an element-wise multiplication and $\mathbf{g}_s\in[0, 1]$ is a gating vector estimated from $w_{t-\ell}$ using an MLP.

\textbf{Spatial Factors ($F$):}
As with the temporal latents, we assume a diagonal Gaussian distribution for spatial factors parameterized with an MLP as $p_\theta(F|z) = \mathcal{N}\big(\mu_\theta^F(z), \Sigma_\theta^F(z)\big)$,
where $z$ itself is sampled from a normal distribution: $z \sim \mathcal{N}(0, I)$. Introducing $z$, as a low dimensional spatial embedding, encourages the estimation of a multimodal distribution among spatial factors. Namely, marginalizing $p(F,z)=p(F|z)p(z)$ over $z$ leads to a Gaussian-mixture prior over $K$ factors in $F$ (given the \emph{nonlinear} mapping from $z$ that parameterizes the Gaussian $p(F|z)$). Whereas a matrix Normal prior on $F$, as in \citet{sun2019bayesian}, naively assumes that $f_{1:K}$ have a unimodal distribution and uncorrelated elements, DSARF is able to encode such correlations by jointly estimating the factors from $z$.

\textbf{Expected Log Likelihood:} Finally, having the temporal weights and spatial factors, we can recover the data by incorporating our initial factorization assumption from \eqnref{fac}:
{\small
\begin{equation}
    X_n \sim p_\theta(X_n|W_n,F) =
    \mathcal{N}\Big(\big[w_{n,1}, \cdots, w_{n,T}\big]^\top F, \sigma_0^2\Big),\nonumber
\end{equation}
}where $\sigma_0$ is a hyperparameter for observation noise.

\textbf{Variational Parameters:} We introduce trainable variational parameters, $\phi$, as mean and diagonal covariance of a Gaussian distribution for each data point to define a fully factorized variational distribution on the latents:
{\small
\begin{equation}
    q(z;\phi^{z}),\: q(F;\phi^F),\:\Big\{q(w_{n,t};\phi^w_{n,t})\Big\}_{n=1,\,t=-\ell}^{N,\,T}\nonumber
\end{equation}
}We approximate variational parameters for discrete latents, $\big\{q(s_{n,t};\phi^s_{n,t})\big\}_{n=1,\,t=1}^{N,\,T}$, with posteriors $\big\{p(s_{n,t}|w_{n,t})\big\}_{n=1,\,t=1}^{N,\,T}$ to compensate information loss induced by the mean-field approximation:
{\small
\begin{align}
    &q(s_{t};\phi^s_{t})\simeq p(s_{t}|w_{t})=\nonumber\\
    &\frac{\mathbb{E}_{q(s_{t-1})q(w_{t-\ell})}\Big[p(s_{t}|s_{t-1})p(w_{t}|w_{t-\ell},s_{t})\Big]}{\sum_{s=1}^S\mathbb{E}_{q(s_{t-1})q(w_{t-\ell})}\Big[p(s_{t}=s|s_{t-1})p(w_{t}|w_{t-\ell},s_{t}=s)\Big]}\hspace{-.2em}\label{eqn:state_posterior}
\end{align}
}This approximation has a two-fold advantage: (1) spares the model additional trainable parameters for the variational distribution, and (2) further couples together the generative and variational parameters of \emph{discrete} and \emph{continuous} latents, and together with \eqnref{state_transition} resolve the \emph{open loop} issue mentioned in \citet{linderman2017bayesian} for these switching models as follows. The posterior on discrete state $s_t$ is informed about the current value of the continuous latent $w_t$ through \eqnref{state_posterior}: $q(s_t;\mathlarger{\boldsymbol{\pi}}_{s_t}) \simeq p_\theta(s_t|w_t)\propto p_\theta(w_t|s_t)p_\theta(s_t)$, $w_t\sim q_\phi(w_t)$, hence $\mathbf{\pi}_{s_t}$ is a function of $w_t$, i.e., $\mathlarger{\boldsymbol{\pi}}_{s_t}=f(w_t;\theta).p_\theta(s_t)$. This is then propagated to the generative model through \eqnref{state_transition}: 
{\small
\begin{align}
p_\theta(s_{t+1}|s_t) = \text{Cat}(\mathbf{\Phi_\theta} \mathlarger{\boldsymbol{\pi}}_{s_t}) = \text{Cat}\left(\mathbf{\Phi_\theta}\,f(w_t;\theta).p_\theta(s_t)\right).\nonumber
\end{align}
}The latter modulates the prior on latent state, $p(s_{t+1})$, with $w_t$, whereas the rSLDS models explicitly allow the discrete switches to depend on the continuous latents.
\synthfig

\textbf{Training DSARF:}
We compute the Monte-Carlo estimate of the gradient of ELBO in \eqnref{elbo_terms} with respect to generative, $\theta$, and variational, $\phi$, parameters using a re-parameterized sample, \citep{kingma2013auto}, from the posterior of continuous latents, $\{W, z, F\}$. For the discrete latent, $\mathcal{S}$, we compute the expectations over $q_{\phi}(s_{n,t})$ by summing over the $S$ possible states, hence no explicit sampling is performed, i.e, for each data point {\small$\boldsymbol{\mathcal{L}_{t}^{\mathcal{W}}}$} would be:
{\small
\begin{align*}
    -\sum_{s=1}^S q(s_{t}=s)\mathbb{E}_{q(w_{t-\ell})} \Big[\text{KL}\big(q(w_{t})\|p\big(w_{t}|w_{t-\ell},s_{t}=s)\big)\Big]\nonumber
\end{align*}
}This explicitly regularizes the $S$ nonlinear auto-regressive priors based on
their corresponding weighting.
We can analytically calculate the Kullback-Leibler (KL) divergence terms of ELBO for both multivariate Gaussian and categorical distributions, which leads to lower variance gradient estimates and faster training as compared to e.g., noisy Monte Carlo estimates often used in literature. 

\textbf{Implementation Details:}
We implemented DSARF in PyTorch v1.3 \citep{paszke2017automatic}, and used the Adam optimizer \citep{kingma2014adam} with learning rate of $0.01$ for training. We initialized all the parameters randomly, and adopted a linear KL annealing schedule, \citep{bowman2015generating}, to increase from $0.01$ to $1$ over the course of $100$ epochs. DSARF has $O($NK$\times($T+D$))$ variational parameters and $O($S$^2+$S$|\ell|$K$^2)$ parameters for the temporal generative model.
We learned and tested all the models on an Intel Core i7 CPU@3.7GHz with 8 GB of RAM. Per-epoch training time varied from $30$ms in smaller datasets to $1.2$ s in larger experiments and $500$ epochs sufficed for most experiments.

\textbf{Long- \& Short-Term Prediction:}
We evaluated the performance of DSARF for both long- and short-term prediction tasks by adopting a rolling prediction scheme \citep{yu2016temporal,chen2019missing}. For long-term prediction, we predict the test set sequentially using the generative model and spatial factors learned on the train set. 
For short-term prediction, we predict the next time point on the test set using the generative model and spatial factors learned on the train set: $\hat{X}_{t+1}=\hat{w}_{t+1}^\top F$, where $\hat{w}_{t+1}\sim p(\hat{w}_{t+1}|{w}_{t+1-\ell}, \hat{s}_{t+1})$, and $\hat{s}_{t+1}\sim p(\hat{s}_{t+1}|s_t)$. We then run inference on $X_{t+1}$, the actual observation at $t+1$ (if not missing), to obtain $w_{t+1}$ and $s_{t+1}$, and add them to the historical data for prediction of the next time point $\hat{X}_{t+2}$ in the same way. We repeat these steps to make short-term predictions in a rolling manner across a test set. We keep the generative model and spatial factors fixed during the entire prediction. We report normalized root-mean-square error (NRMSE\%) for both long- and short-term predictions (see \suppref{nrmse}). The test set NRMSE\% we report for short-term predictions is related to the expected \emph{negative test-set log-likelihood} for our case of Gaussian distributions (with a multiplicative/additive constant), hence it is used for evaluating the predictive generative models.

\section{Experimental Evaluation}
\label{sec:experimental}
\longfig
We evaluated the performance of DSARF in modeling the temporal dynamics and discovering the underlying temporal states by conducting a number of controlled synthetic experiments, followed by a comprehensive real-world data assessment covering a wide range of application areas. To this end, we compared the predictive performance of DSARF against two established Bayesian switching state-space models, rSLDS \citep{nassar2019tree} and SLDS \citep{fox2009nonparametric}, three state-of-the-art dynamical matrix factorization methods, BTMF \citep{sun2019bayesian}, TRMF \citep{yu2016temporal} and its Bayesian extension (B-TRMF), a deep state-space model, RKN \citep{becker2019recurrent}, and a deep neural network-based time series forecasting method, LSTNet \citep{lai2018modeling}, which employs vector auto-regression, and allows long-term forecasting, in terms of short- and/or long-term prediction tasks throughout the experiments.
\subsection{Model Evaluation using Synthetic Data}
\label{sec:synthetic_exp}
\textbf{Toy Example:} Inspired by \citet{ghahramani1996switching}, we generated $N=200$ spatio-temporal sequences, each with $T=200$ time points and $D=10$ spatial dimensions from $K=2$ shared factors according to a simple nonlinear dynamical model in $W$ which switched between two temporal models as follows:
{\small
\begin{align}
    &w_{t|s_t = 0} = 0.9\, w_{t-1} + \tanh(0.5w_{t-2}) + 3\sin(w_{t-3})+\epsilon^0_{t}\nonumber\\
    &w_{t|s_t = 1} = 0.9\, w_{t-1} + \tanh(0.2w_{t-2}) + \sin(w_{t-3})+\epsilon^1_{t}\nonumber\\
    &X_n = [w_{1}, \cdots, w_{t}]_n\, \text{F} + \nu_n \quad \nu_n\sim\mathcal{N}(0,0.1\,\mathbf{I}),\; \text{F} \sim \mathcal{U}(-1,1)\nonumber
\end{align}
}where $w_{t}\in\mathbb{R}^2$, $\text{F}\in\mathbb{R}^{2\times10}$ and $\epsilon_{t}\sim\mathcal{N}(0,\mathbf{I})$, and the switch state $s_{t}$ was chosen using priors $\pi_1 = \pi_2 = 1/2$ and transition probabilities $\Phi_{11} = \Phi_{22} =0.95$; $\Phi_{12} = \Phi_{21} =0.05$. We picked 10 sequences for test, and trained DSARF on the rest with lag set $\ell=\{1,2,3\}$ for $200$ epochs. We recovered temporal states on the entire dataset with an accuracy of $79.63\%\pm 3.89$ compared to $61.74\%\pm 9.13$ and $52.01\%\pm 11.19$ for rSLDS and SLDS respectively. We predicted the test set in short-term with NRMSE of $13.81$\% while rSLDS and SLDS only achieved $78.45$\% and $98.01$\% respectively. rSLDS and SLDS apparently failed in modeling the higher order temporal dependencies in this synthetic data. We have visualized short-term predictions along with recovered states for three test sequences in \figref{synthetic}a.
\\
\hphantom\quad\textbf{Lorenz Attractor:}
We applied DSARF to simulated data from a canonical nonlinear dynamical system, the \emph{Lorenz attractor}, whose nonlinear dynamics are given by: 
{\small
\begin{equation}
    \frac{d\mathbf{w}}{dt} = \begin{bmatrix}\alpha(\text{w}_2-\text{w}_1)\\\text{w}_1(\beta-\text{w}_1)-\text{w}_2\\\text{w}_1\text{w}_2-\gamma \text{w}_3\end{bmatrix}\nonumber
\end{equation}
}Though nonlinear and chaotic, we see that the Lorenz attractor roughly traces out ellipses in two opposing planes (see \figref{synthetic}b top-right). We simulated $T=2000$ time points and left the second half for test. Rather than directly observing the states, $\text{w}_{1:T}$, we projected them into a $D = 10$ dimensional space: $X=W^\top F$. Fitting DSARF, we found that the model separates these two planes into two distinct states (accuracy of $92.90\%$), each with rotational dynamics as depicted in \figref{synthetic}b bottom-left. SLDS failed in detecting the true states, while rSLDS performed close in terms of state estimation (see \figref{synthetic}b). However, DSARF predicted the test set in short-term with NRMSE of $0.88\%$ compared to $1.14\%$ and $2.70\%$ for rSLDS and SLDS respectively. Note that the latent states are only identifiable up to invertible transformation.
\\
\hphantom\quad\textbf{Double Pendulum:}
A double pendulum (a pendulum with another pendulum attached to its end) is another simple nonlinear physical system that exhibits rich dynamic behavior with a strong sensitivity to initial conditions. The motion of a double pendulum is governed by a set of coupled second-order ordinary differential equations and is chaotic \citep{levien1993double}:
{\small
\begin{align}
    &2\ddot{\theta}_1+\ddot{\theta}_2\cos(\theta_1-\theta_2)+\dot{\theta}_2^2\sin(\theta_1-\theta_2)+2g\sin(\theta_1)=0\nonumber\\
    &\ddot{\theta}_2+\ddot{\theta}_1\cos(\theta_1-\theta_2)+\dot{\theta}_1^2\sin(\theta_1-\theta_2)+g\sin(\theta_2)=0,
    \nonumber
\end{align}
}where $\theta_1$ and $\theta_2$ are the deflection angles of the pendulums, and $g$ is the gravitational acceleration. We simulated the system for $T=20,000$ time points and recorded the locations of the two pendulums. We observed these locations through a linear projection with $D=10$ just like the previous experiments. We kept the last $400$ time points for test, and fit DSARF with $S=3$ once with lag set $\ell=\{1,2\}$ and another time with $\ell=\{1\}$ on the train set. DSARF with $\ell=\{1,2\}$ predicted the test set in short-term with NRMSE of $4.38\%$, while DSARF with $\ell=\{1\}$ achieved $9.79\%$, and rSLDS and SLDS achieved $10.42\%$ and $15.53\%$ respectively. This is expected as the second derivatives of location (i.e., acceleration) appear in the Euler-Lagrange differential equation for double pendulum. We have visualized short-term predictions of the test set along with inferred states and dynamics in \figref{synthetic}c. While this system could potentially be segmented to more states, we observed that for $S=3$ the dynamical trajectory is roughly segmented along the deflection angle of the first pendulum. Increasing $S=10$ would further improve test set prediction error to $4.16\%$.
\ErrTable
\DataTable
\subsection{Model Evaluation using Real-World Data}
We give a brief description of each real-world data as well as the train/test splits we used for our evaluation in \tblref{DataTable}. In the following paragraphs, we describe our quantitative comparison results summarized in \tblref{ErrTable}.

\textbf{Short-term prediction results:} For the sleep Apnea dataset, we observed that DSARF is able to segment the respiration signal for both train and test sets into instances of apnea (small variability in chest volume, followed by bursts) with an accuracy of $86\%$ on the test set compared to $71\%$ for SLDS, while rSLDS completely fails on this task as depicted in \figref{long}j. In addition, DSARF predicts the test set in short-term with NRMSE of $23.86\%$ outperforming all the other baselines (see \tblref{ErrTable}. RKN with $27.13\%$ is the second best). For the other datasets, as summarized in the short-term section of \tblref{ErrTable}, DSARF outperforms in short-term prediction of test sets in Birmangham, Hangzhou, Seattle, Bat flight and precipitation datasets, while closely following the state-of-the-art in Google flu and Dengue datasets (where TRMF and B-TRMF perform the best respectively). LSTNet and rSLDS perform better in Guangzhou and PST datasets respectively, while rSLDS follows DSARF closely in Seattle and Precipitation datasets. We reported the results with (w/) and without (w/o) the switching feature for DSARF, rSLDS and SLDS to explore the impact of these switching latents. We observed that the predictions for Birmingham, Bat, PST, Precipitation and Google flu datasets improved when the switching feature was employed in DSARF. We used lag set $\ell=\{1,2\}$ for DSARF on all short-term prediction experiments (set accordingly for BTMF, TRMF, B-TRMF, and LSTNet). Sample short-term predictions of test set for precipitation and bat flight data are depicted in \figref{long}h, i respectively.

\textbf{Long-term prediction results:} We excluded the Bat flight, Precipitation, and Apnea datasets from long-term prediction task as these datasets hardly show periodic behaviours and/or are chaotic, e.g., in precipitation data \citep{buizza2002chaos}. On the other hand, we see some extent of long-term recurrence, e.g., in traffic data and seasonal diseases spread, over calendar dates (days, weeks, seasons, etc.). For this reason, we used the lag set $\ell=\{1, 2, 3, T_0, T_0+1, T_0+2, 7T_0, 7T_0+1, 7T_0+2\}$ for traffic datasets as in \citet{sun2019bayesian} (where $T_0$ is the time points per day), $\ell=\{1,2, 52, 52+1, 2\times52, 2\times52+1\}$ weeks for Google flu and Dengue datasets and $\ell=\{1, 2, 12, 12+1, 6\times12, 6\times12+1\}$ months for the PST dataset. We also excluded SLDS, rSLDS and RKN from this comparison as these baselines do not allow for long historical conditioning, hence are intractable for long-term prediction and diverge very fast. As summarized in the long-term section of \tblref{ErrTable}, DSARF outperforms in long-term prediction of the test sets in Birmingham, Hangzhou, Seattle, PST, Google flu and Dengue datasets, while closely follows BTMF on the Guangzhou dataset. We have visualized sample long-term predictions of test set (one spatial dimension per dataset) along with prediction uncertainty and ground-truth values in \figref{long}a-g. Note that part of the error is sourced from the sparse factorization.

\textbf{Spatial generative model:} DSARF resulted in spatial factors with higher test-set log-likelihood in all of our real-data experiments when compared to a widely used matrix Normal prior, with $-1.02$ nats versus $-1.37$ nats (on average) respectively (see \suppref{spatial_generative} and \tblref{spatialTable} for details).
\section{Conclusion}
We introduced deep switching auto-regressive factorization (DSARF) in a Bayesian framework. Our method extends switching linear dynamical system models and Bayesian dynamical matrix factorization methods by employing a non-linear vector auto-regressive latent model switched by a Markovian chain of discrete latents to capture higher-order multimodal latent dependencies. This expands prediction horizon and improves long- and short-term forecasting as demonstrated by our extensive synthetic and real data experiments. DSARF proves scalable to high-dimensional data due to the incorporation of factorization framework, is tractable on missing data, provides uncertainty measures for estimations, and lends itself to an efficient inference algorithm.
\label{sec:conclu}

\section*{Ethics Statement}
The model we proposed in this work is a step toward better understanding of high dimensional time series data that appear in a variety of real-world settings. Analysing and more importantly \emph{forecasting} these times series naturally embrace a very broad range of applications from healthcare management, disease spread prediction and infection diagnosis to traffic control and weather and  financial forecasting, which are where we see the potential for a broader impact. Although, these time series data often show long- and short-term recurring patterns, they occasionally exhibit sophisticated behaviours or are chaotic. Subsequently, we need appropriate tools and legitimate assumptions for analysing them. While we understand that, as George Box wrote in \citet{box1979robustness}, ``all models are wrong but some are useful,'' we hope that in a wide range of applications with proper assumptions and prior knowledge, our DSARF model be useful in providing a means to analyzing high dimensional spatio-temporal data. 

{\small
\bibliography{paper}

\begin{thebibliography}{63}
\providecommand{\natexlab}[1]{#1}
\providecommand{\url}[1]{\texttt{#1}}
\providecommand{\urlprefix}{URL }
\expandafter\ifx\csname urlstyle\endcsname\relax
  \providecommand{\doi}[1]{doi:\discretionary{}{}{}#1}\else
  \providecommand{\doi}{doi:\discretionary{}{}{}\begingroup
  \urlstyle{rm}\Url}\fi

\bibitem[{IRI(2002)}]{IRI/LDEO}
 2002.
\newblock Pacific Ocean Temperature Dataset.
\newblock \url{http://iridl.ldeo.columbia.edu/}.

\bibitem[{SIL(2015)}]{SILD}
 2015.
\newblock Seattle Inductive Loop Detector Dataset V.1.
\newblock \url{https://github.com/zhiyongc/Seattle-Loop-Data}.

\bibitem[{Den(2016)}]{Dengue}
 2016.
\newblock Google Dengue Trends Data.
\newblock \url{https://www.google.org/flutrends/about/}.

\bibitem[{Flu(2016)}]{Flu}
 2016.
\newblock Google Flu Trends Data.
\newblock \url{https://www.google.org/flutrends/about/}.

\bibitem[{B(2016)}]{B}
 2016.
\newblock Parking Birmingham Data Set.
\newblock \url{https://data.birmingham.gov.uk/dataset/birmingham-parking}.

\bibitem[{UTS(2016)}]{UTSD}
 2016.
\newblock Urban Traffic Speed Dataset of Guangzhou, China.
\newblock \url{https://doi.org/10.5281/zenodo.1205229}.

\bibitem[{NOA(2017)}]{NOAA}
 2017.
\newblock Colorado Precipitation Dataset.
\newblock \url{https://www.ncdc.noaa.gov/}.

\bibitem[{HIP(2019)}]{HIPF}
 2019.
\newblock Hangzhou Incoming Passenger Flow.
\newblock \url{https://tianchi.aliyun.com/competition/entrance/231708/}.

\bibitem[{Ackerson and Fu(1970)}]{ackerson1970state}
Ackerson, G.; and Fu, K. 1970.
\newblock On state estimation in switching environments.
\newblock \emph{IEEE transactions on automatic control} 15(1): 10--17.

\bibitem[{Bahadori, Yu, and Liu(2014)}]{bahadori2014fast}
Bahadori, M.~T.; Yu, Q.~R.; and Liu, Y. 2014.
\newblock Fast multivariate spatio-temporal analysis via low rank tensor
  learning.
\newblock In \emph{Advances in neural information processing systems},
  3491--3499.

\bibitem[{Becker et~al.(2019)Becker, Pandya, Gebhardt, Zhao, Taylor, and
  Neumann}]{becker2019recurrent}
Becker, P.; Pandya, H.; Gebhardt, G.; Zhao, C.; Taylor, C.~J.; and Neumann, G.
  2019.
\newblock Recurrent Kalman Networks: Factorized Inference in High-Dimensional
  Deep Feature Spaces.
\newblock In \emph{International Conference on Machine Learning}, 544--552.

\bibitem[{Becker-Ehmck, Peters, and van~der Smagt(2019)}]{becker2019switching}
Becker-Ehmck, P.; Peters, J.; and van~der Smagt, P. 2019.
\newblock Switching Linear Dynamics for Variational Bayes Filtering .

\bibitem[{Bergou et~al.(2015)Bergou, Swartz, Vejdani, Riskin, Reimnitz, Taubin,
  and Breuer}]{bergou2015falling}
Bergou, A.~J.; Swartz, S.~M.; Vejdani, H.; Riskin, D.~K.; Reimnitz, L.; Taubin,
  G.; and Breuer, K.~S. 2015.
\newblock Falling with style: bats perform complex aerial rotations by
  adjusting wing inertia.
\newblock \emph{PLoS Biol} 13(11): e1002297.

\bibitem[{Bowman et~al.(2016)Bowman, Vilnis, Vinyals, Dai, Jozefowicz, and
  Bengio}]{bowman2015generating}
Bowman, S.~R.; Vilnis, L.; Vinyals, O.; Dai, A.~M.; Jozefowicz, R.; and Bengio,
  S. 2016.
\newblock Generating Sentences from a Continuous Space.
\newblock \emph{CoNLL 2016} 10.

\bibitem[{Box(1979)}]{box1979robustness}
Box, G.~E. 1979.
\newblock Robustness in the strategy of scientific model building.
\newblock In \emph{Robustness in statistics}, 201--236.

\bibitem[{Buizza(2002)}]{buizza2002chaos}
Buizza, R. 2002.
\newblock Chaos and weather prediction January 2000.
\newblock \emph{European Centre for Medium-Range Weather Meteorological
  Training Course Lecture Series ECMWF} .

\bibitem[{Cai et~al.(2015)Cai, Tong, Fan, Ji, and He}]{cai2015facets}
Cai, Y.; Tong, H.; Fan, W.; Ji, P.; and He, Q. 2015.
\newblock Facets: Fast comprehensive mining of coevolving high-order time
  series.
\newblock In \emph{ACM SIGKDD International Conference on Knowledge Discovery
  and Data Mining}, 79--88.

\bibitem[{Chang and Athans(1978)}]{chang1978state}
Chang, C.-B.; and Athans, M. 1978.
\newblock State estimation for discrete systems with switching parameters.
\newblock \emph{IEEE Transactions on Aerospace and Electronic Systems} (3):
  418--425.

\bibitem[{Chang et~al.(2018)Chang, Sun, Wu, and Lin}]{chang2018memory}
Chang, Y.-Y.; Sun, F.-Y.; Wu, Y.-H.; and Lin, S.-D. 2018.
\newblock A memory-network based solution for multivariate time-series
  forecasting.
\newblock \emph{arXiv preprint arXiv:1809.02105} .

\bibitem[{Charlin et~al.(2015)Charlin, Ranganath, McInerney, and
  Blei}]{charlin2015dynamic}
Charlin, L.; Ranganath, R.; McInerney, J.; and Blei, D.~M. 2015.
\newblock Dynamic poisson factorization.
\newblock In \emph{Proceedings of the 9th ACM Conference on Recommender
  Systems}, 155--162.

\bibitem[{Che et~al.(2018)Che, Purushotham, Cho, Sontag, and
  Liu}]{che2018recurrent}
Che, Z.; Purushotham, S.; Cho, K.; Sontag, D.; and Liu, Y. 2018.
\newblock Recurrent neural networks for multivariate time series with missing
  values.
\newblock \emph{Scientific reports} 8(1): 1--12.

\bibitem[{Chen et~al.(2019)Chen, He, Chen, Lu, and Wang}]{chen2019missing}
Chen, X.; He, Z.; Chen, Y.; Lu, Y.; and Wang, J. 2019.
\newblock Missing traffic data imputation and pattern discovery with a Bayesian
  augmented tensor factorization model.
\newblock \emph{Transportation Research Part C: Emerging Technologies} 104:
  66--77.

\bibitem[{Farnoosh et~al.(2020)Farnoosh, Rezaei, Sennesh, Khan, Dy, Satpute,
  Hutchinson, van~de Meent, and Ostadabbas}]{farnoosh2020deep}
Farnoosh, A.; Rezaei, B.; Sennesh, E.~Z.; Khan, Z.; Dy, J.; Satpute, A.;
  Hutchinson, J.~B.; van~de Meent, J.-W.; and Ostadabbas, S. 2020.
\newblock Deep Markov Spatio-Temporal Factorization.
\newblock \emph{arXiv preprint arXiv:2003.09779} .

\bibitem[{Fox et~al.(2009)Fox, Sudderth, Jordan, and
  Willsky}]{fox2009nonparametric}
Fox, E.; Sudderth, E.~B.; Jordan, M.~I.; and Willsky, A.~S. 2009.
\newblock Nonparametric Bayesian learning of switching linear dynamical
  systems.
\newblock In \emph{Advances in neural information processing systems},
  457--464.

\bibitem[{Fraccaro et~al.(2017)Fraccaro, Kamronn, Paquet, and
  Winther}]{fraccaro2017disentangled}
Fraccaro, M.; Kamronn, S.; Paquet, U.; and Winther, O. 2017.
\newblock A disentangled recognition and nonlinear dynamics model for
  unsupervised learning.
\newblock In \emph{Advances in Neural Information Processing Systems},
  3601--3610.

\bibitem[{Ghahramani and Hinton(1996)}]{ghahramani1996switching}
Ghahramani, Z.; and Hinton, G.~E. 1996.
\newblock Switching state-space models.
\newblock Technical report, Citeseer.

\bibitem[{Ghazi et~al.(2018)Ghazi, Nielsen, Pai, Cardoso, Modat, Ourselin, and
  Sorensen}]{ghazi2018robust}
Ghazi, M.~M.; Nielsen, M.; Pai, A. S.~U.; Cardoso, M.~J.; Modat, M.; Ourselin,
  S.; and Sorensen, L. 2018.
\newblock Robust training of recurrent neural networks to handle missing data
  for disease progression modeling.
\newblock In \emph{1st Conference on Medical Imaging with Deep Learning (MIDL
  2018)}.

\bibitem[{Goldberger et~al.(2000)Goldberger, Amaral, Glass, Hausdorff, Ivanov,
  Mark, Mietus, Moody, Peng, and Stanley}]{goldberger2000physiobank}
Goldberger, A.~L.; Amaral, L.~A.; Glass, L.; Hausdorff, J.~M.; Ivanov, P.~C.;
  Mark, R.~G.; Mietus, J.~E.; Moody, G.~B.; Peng, C.-K.; and Stanley, H.~E.
  2000.
\newblock PhysioBank, PhysioToolkit, and PhysioNet: components of a new
  research resource for complex physiologic signals.
\newblock \emph{circulation} 101(23): e215--e220.

\bibitem[{Hamilton(1990)}]{hamilton1990analysis}
Hamilton, J.~D. 1990.
\newblock Analysis of time series subject to changes in regime.
\newblock \emph{Journal of econometrics} 45(1-2): 39--70.

\bibitem[{Hoffman et~al.(2013)Hoffman, Blei, Wang, and
  Paisley}]{hoffman2013stochastic}
Hoffman, M.~D.; Blei, D.~M.; Wang, C.; and Paisley, J. 2013.
\newblock Stochastic variational inference.
\newblock \emph{The Journal of Machine Learning Research} 14(1): 1303--1347.

\bibitem[{Jing et~al.(2018)Jing, Su, Jin, and Zhang}]{jing2018high}
Jing, P.; Su, Y.; Jin, X.; and Zhang, C. 2018.
\newblock High-order temporal correlation model learning for time-series
  prediction.
\newblock \emph{IEEE transactions on cybernetics} 49(6): 2385--2397.

\bibitem[{Juloski, Weiland, and Heemels(2005)}]{juloski2005bayesian}
Juloski, A.~L.; Weiland, S.; and Heemels, W. 2005.
\newblock A Bayesian approach to identification of hybrid systems.
\newblock \emph{IEEE Transactions on Automatic Control} 50(10): 1520--1533.

\bibitem[{Karl et~al.(2017)Karl, Soelch, Bayer, and van~der
  Smagt}]{karl2017deep}
Karl, M.; Soelch, M.; Bayer, J.; and van~der Smagt, P. 2017.
\newblock Deep variational bayes filters: Unsupervised learning of state space
  models from raw data.
\newblock \emph{stat} 1050: 3.

\bibitem[{Kingma and Ba(2014)}]{kingma2014adam}
Kingma, D.~P.; and Ba, J. 2014.
\newblock Adam: A method for stochastic optimization.
\newblock \emph{arXiv preprint arXiv:1412.6980} .

\bibitem[{Kingma and Welling(2014)}]{kingma2013auto}
Kingma, D.~P.; and Welling, M. 2014.
\newblock Auto-Encoding Variational Bayes.
\newblock \emph{stat} 1050: 1.

\bibitem[{Krishnan, Shalit, and Sontag(2015)}]{krishnan2015deep}
Krishnan, R.~G.; Shalit, U.; and Sontag, D. 2015.
\newblock Deep Kalman Filters.
\newblock \emph{stat} 1050: 25.

\bibitem[{Krishnan, Shalit, and Sontag(2017)}]{krishnan2017structured}
Krishnan, R.~G.; Shalit, U.; and Sontag, D. 2017.
\newblock Structured inference networks for nonlinear state space models.
\newblock In \emph{Thirty-First AAAI Conference on Artificial Intelligence}.

\bibitem[{Lai et~al.(2018)Lai, Chang, Yang, and Liu}]{lai2018modeling}
Lai, G.; Chang, W.-C.; Yang, Y.; and Liu, H. 2018.
\newblock Modeling long-and short-term temporal patterns with deep neural
  networks.
\newblock In \emph{ACM SIGIR Conference on Research \& Development in
  Information Retrieval}, 95--104.

\bibitem[{Levien and Tan(1993)}]{levien1993double}
Levien, R.; and Tan, S. 1993.
\newblock Double pendulum: An experiment in chaos.
\newblock \emph{American Journal of Physics} 61(11): 1038--1044.

\bibitem[{Li et~al.(2019)Li, Jin, Xuan, Zhou, Chen, Wang, and
  Yan}]{li2019enhancing}
Li, S.; Jin, X.; Xuan, Y.; Zhou, X.; Chen, W.; Wang, Y.-X.; and Yan, X. 2019.
\newblock Enhancing the locality and breaking the memory bottleneck of
  transformer on time series forecasting.
\newblock In \emph{Advances in Neural Information Processing Systems}.

\bibitem[{Linderman et~al.(2017)Linderman, Johnson, Miller, Adams, Blei, and
  Paninski}]{linderman2017bayesian}
Linderman, S.; Johnson, M.; Miller, A.; Adams, R.; Blei, D.; and Paninski, L.
  2017.
\newblock Bayesian learning and inference in recurrent switching linear
  dynamical systems.
\newblock In \emph{Artificial Intelligence and Statistics}, 914--922.

\bibitem[{Mattei and Frellsen(2019)}]{mattei2019miwae}
Mattei, P.-A.; and Frellsen, J. 2019.
\newblock MIWAE: Deep generative modelling and imputation of incomplete data
  sets.
\newblock In \emph{International Conference on Machine Learning}, 4413--4423.

\bibitem[{Murphy(1998)}]{murphy1998switching}
Murphy, K.~P. 1998.
\newblock Switching kalman filters .

\bibitem[{Nassar et~al.(2019)Nassar, Linderman, Bugallo, and
  Park}]{nassar2019tree}
Nassar, J.; Linderman, S.; Bugallo, M.; and Park, I. 2019.
\newblock Tree-Structured Recurrent Switching Linear Dynamical Systems for
  Multi-Scale Modeling.
\newblock In \emph{International Conference on Learning Representations
  (ICLR)}.

\bibitem[{Paoletti et~al.(2007)Paoletti, Juloski, Ferrari-Trecate, and
  Vidal}]{paoletti2007identification}
Paoletti, S.; Juloski, A.~L.; Ferrari-Trecate, G.; and Vidal, R. 2007.
\newblock Identification of hybrid systems a tutorial.
\newblock \emph{European journal of control} 13(2-3): 242--260.

\bibitem[{Paszke et~al.(2017)Paszke, Gross, Chintala, Chanan, Yang, DeVito,
  Lin, Desmaison, Antiga, and Lerer}]{paszke2017automatic}
Paszke, A.; Gross, S.; Chintala, S.; Chanan, G.; Yang, E.; DeVito, Z.; Lin, Z.;
  Desmaison, A.; Antiga, L.; and Lerer, A. 2017.
\newblock Automatic differentiation in PyTorch .

\bibitem[{Ranganath et~al.(2013)Ranganath, Wang, David, and
  Xing}]{ranganath2013adaptive}
Ranganath, R.; Wang, C.; David, B.; and Xing, E. 2013.
\newblock An adaptive learning rate for stochastic variational inference.
\newblock In \emph{International Conference on Machine Learning}, 298--306.

\bibitem[{Rangapuram et~al.(2018)Rangapuram, Seeger, Gasthaus, Stella, Wang,
  and Januschowski}]{rangapuram2018deep}
Rangapuram, S.~S.; Seeger, M.~W.; Gasthaus, J.; Stella, L.; Wang, Y.; and
  Januschowski, T. 2018.
\newblock Deep state space models for time series forecasting.
\newblock In \emph{Advances in neural information processing systems},
  7785--7794.

\bibitem[{Rezende and Mohamed(2015)}]{rezende2015variational}
Rezende, D.~J.; and Mohamed, S. 2015.
\newblock Variational inference with normalizing flows.
\newblock In \emph{International Conference on Machine Learning}, 1530--1538.

\bibitem[{Rigney(1994)}]{rigney1994multichannel}
Rigney, D.~R. 1994.
\newblock Multichannel physiological data description and analysis.
\newblock \emph{Time Series Prediction} .

\bibitem[{Riskin et~al.(2008)Riskin, Willis, Iriarte-D{\'\i}az, Hedrick,
  Kostandov, Chen, Laidlaw, Breuer, and Swartz}]{riskin2008quantifying}
Riskin, D.~K.; Willis, D.~J.; Iriarte-D{\'\i}az, J.; Hedrick, T.~L.; Kostandov,
  M.; Chen, J.; Laidlaw, D.~H.; Breuer, K.~S.; and Swartz, S.~M. 2008.
\newblock Quantifying the complexity of bat wing kinematics.
\newblock \emph{Journal of theoretical biology} 254(3): 604--615.

\bibitem[{Rogers, Li, and Russell(2013)}]{rogers2013multilinear}
Rogers, M.; Li, L.; and Russell, S.~J. 2013.
\newblock Multilinear dynamical systems for tensor time series.
\newblock In \emph{Advances in Neural Information Processing Systems},
  2634--2642.

\bibitem[{Salakhutdinov and Mnih(2008)}]{salakhutdinov2008bayesian}
Salakhutdinov, R.; and Mnih, A. 2008.
\newblock Bayesian probabilistic matrix factorization using Markov chain Monte
  Carlo.
\newblock In \emph{Proceedings of the 25th international conference on Machine
  learning}, 880--887.

\bibitem[{Salinas et~al.(2020)Salinas, Flunkert, Gasthaus, and
  Januschowski}]{salinas2020deepar}
Salinas, D.; Flunkert, V.; Gasthaus, J.; and Januschowski, T. 2020.
\newblock DeepAR: Probabilistic forecasting with autoregressive recurrent
  networks.
\newblock \emph{International Journal of Forecasting} 36(3): 1181--1191.

\bibitem[{Sen, Yu, and Dhillon(2019)}]{sen2019think}
Sen, R.; Yu, H.-F.; and Dhillon, I.~S. 2019.
\newblock Think globally, act locally: A deep neural network approach to
  high-dimensional time series forecasting.
\newblock In \emph{Advances in Neural Information Processing Systems},
  4837--4846.

\bibitem[{Sontag(1981)}]{sontag1981nonlinear}
Sontag, E. 1981.
\newblock Nonlinear regulation: The piecewise linear approach.
\newblock \emph{IEEE Transactions on automatic control} 26(2): 346--358.

\bibitem[{Sun, Parthasarathy, and Varshney(2014)}]{sun2014collaborative}
Sun, J.~Z.; Parthasarathy, D.; and Varshney, K.~R. 2014.
\newblock Collaborative kalman filtering for dynamic matrix factorization.
\newblock \emph{IEEE Transactions on Signal Processing} 62(14): 3499--3509.

\bibitem[{Sun and Chen(2019)}]{sun2019bayesian}
Sun, L.; and Chen, X. 2019.
\newblock Bayesian Temporal Factorization for Multidimensional Time Series
  Prediction.
\newblock \emph{arXiv preprint arXiv:1910.06366} .

\bibitem[{Takeuchi, Kashima, and Ueda(2017)}]{takeuchi2017autoregressive}
Takeuchi, K.; Kashima, H.; and Ueda, N. 2017.
\newblock Autoregressive tensor factorization for spatio-temporal predictions.
\newblock In \emph{2017 IEEE International Conference on Data Mining (ICDM)},
  1105--1110. IEEE.

\bibitem[{Watter et~al.(2015)Watter, Springenberg, Boedecker, and
  Riedmiller}]{watter2015embed}
Watter, M.; Springenberg, J.; Boedecker, J.; and Riedmiller, M. 2015.
\newblock Embed to control: A locally linear latent dynamics model for control
  from raw images.
\newblock In \emph{Advances in neural information processing systems},
  2746--2754.

\bibitem[{Xiong et~al.(2010)Xiong, Chen, Huang, Schneider, and
  Carbonell}]{xiong2010temporal}
Xiong, L.; Chen, X.; Huang, T.-K.; Schneider, J.; and Carbonell, J.~G. 2010.
\newblock Temporal collaborative filtering with bayesian probabilistic tensor
  factorization.
\newblock In \emph{2010 SIAM international conference on data mining},
  211--222.

\bibitem[{Yu, Rao, and Dhillon(2016)}]{yu2016temporal}
Yu, H.-F.; Rao, N.; and Dhillon, I.~S. 2016.
\newblock Temporal regularized matrix factorization for high-dimensional time
  series prediction.
\newblock In \emph{Advances in neural information processing systems},
  847--855.

\bibitem[{Zhao, Zhang, and Cichocki(2015)}]{zhao2015bayesian}
Zhao, Q.; Zhang, L.; and Cichocki, A. 2015.
\newblock Bayesian CP factorization of incomplete tensors with automatic rank
  determination.
\newblock \emph{IEEE transactions on pattern analysis and machine intelligence}
  37(9): 1751--1763.

\end{thebibliography}
}
\supp
\end{document}